\documentclass[10pt,twocolumn,letterpaper]{article}

\usepackage{cvpr}              
\usepackage{booktabs,multirow,array,colortbl}
\usepackage[table]{xcolor}
\usepackage{pifont}  

\usepackage{tocloft}

\usepackage{times}
\usepackage{helvet}
\usepackage{courier}
\usepackage{xcolor}
\usepackage{natbib}
\usepackage{tcolorbox}
\usepackage{enumitem}
\usepackage{graphicx}
\usepackage{amsmath}

\usepackage{xurl} 

\newcommand{\cmark}{\textcolor{green!40!black}{\ding{52}}} 
\newcommand{\xmark}{\textcolor{red}{\ding{56}}}             

\newcolumntype{P}[1]{>{\centering\arraybackslash}p{#1}} 

\definecolor{cvprblue}{rgb}{0.21,0.49,0.74}
\usepackage[pagebackref,breaklinks,colorlinks,allcolors=cvprblue]{hyperref}

\def\paperID{***} 
\def\confName{CVPR}
\def\confYear{2026}

\title{	
Adaptive Reinforcement for Open-ended Medical Reasoning via Semantic-Guided Reward Collapse Mitigation}

\author{
    Yizhou Liu$^{1,2,*,\S}$ \quad 
    Dingkang Yang$^{1,2,*}$ $\quad$  
    Zizhi Chen$^{1,2}$ \quad
    Minghao Han$^{1,2}$ \\
    Xukun Zhang$^{1}$ \quad
    Keliang Liu$^{1,2}$ \quad
    Jingwei Wei$^{3,\dagger}$ \quad
    Lihua Zhang$^{1,2,\,\dagger}$ \\ 
    \small $^*$equal contribution $\quad$ $^\dagger$corresponding authors $\quad$\\
    $^1$College of Intelligent Robotics and Advanced Manufacturing, Fudan University\\
    $^2$Fysics Intelligence Technologies Co., Ltd. (Fysics AI) \\
    $^3$Institute of Automation, Chinese Academy of Sciences\\
    {\tt\small liuyz25@m.fudan.edu.cn, dicken@fysics.ai,  weijingwei2014@ia.ac.cn, lihuazhang@fudan.edu.cn}
}

\begin{document}
\maketitle
\addtocontents{toc}{\protect\setcounter{tocdepth}{-1}}
\begingroup
\renewcommand{\thefootnote}{\textsection}
\footnotetext{Work done during the collaboration with the Institute of Automation.}
\endgroup

\begin{abstract}
Reinforcement learning (RL) with rule-based reward functions has recently shown great promise in enhancing the reasoning depth and generalization ability of vision-language models (VLMs), while maintaining computational efficiency. In spite of these advances, its adoption in medical imaging remains limited. Current reinforcement fine-tuning (RFT) efforts in this field mainly focus on closed-ended visual question answering (VQA), restricting their applicability to realistic clinical reasoning. However, open-ended medical VQA better mirrors clinical diagnostic workflows but remains underexplored. Although several studies have attempted to bridge the two formats through semantically guided RL, model-driven semantic rewards often suffer from reward collapse, where responses with distinct semantics yield nearly identical scores. To overcome this limitation, we introduce Adaptive Reinforcement for Medical Reasoning (ARMed), a novel RL framework tailored for open-ended medical VQA. ARMed first injects domain expertise through supervised fine-tuning (SFT) on chain-of-thought annotations, followed by reinforcement optimization using textual correctness and adaptive semantic rewards to refine reasoning consistency and factual accuracy. Extensive experiments on six challenging medical VQA benchmarks demonstrate that ARMed substantially improves both accuracy and generalization. These findings underscore the importance of reward discriminability in medical RL and highlight the potential of adaptive semantic rewards for building robust, clinically reliable multimodal reasoning systems.

\end{abstract}    
\section{Introduction}
\begin{figure*}[t]
\centering
\includegraphics[width=\linewidth]{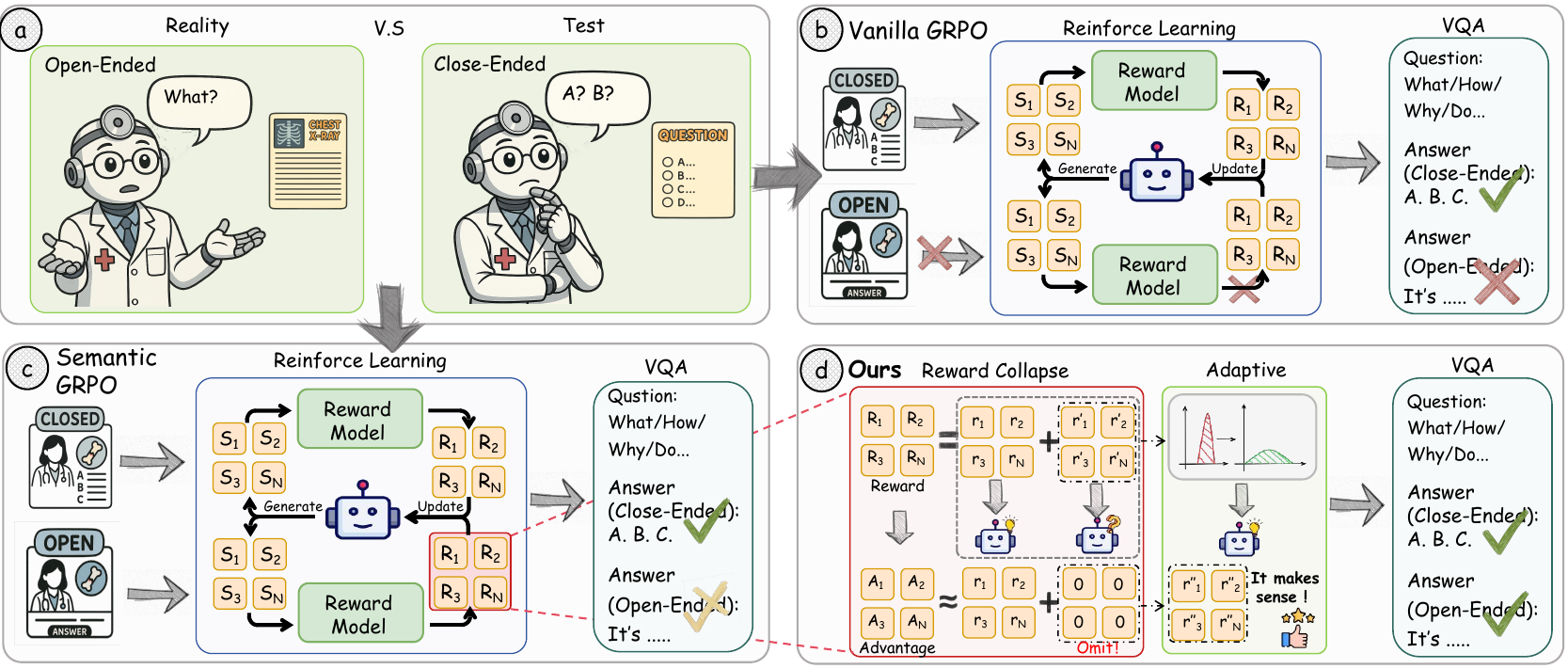} 
\caption{Illustration of the reward collapse issue caused by continuous semantic rewards and comparison among different GRPO variants. The upper part shows the vanilla GRPO trained in closed-ended settings, while real-world tasks are often open-ended. The lower part depicts how semantic-based GRPO can suffer from reward collapse due to the continuity of semantic rewards. Our proposed ARMed method adaptively mitigates this problem, aligning learning between open- and closed-ended scenarios.}
\label{fig:problem}
\vspace{-0.5em}
\end{figure*}

Open-ended medical visual question answering (VQA) mimics how clinicians integrate multimodal information for complex, high-stakes, and real-world diagnostic reasoning. Unlike multiple-choice tasks, it demands highly demanding responses that are both accurate and explanatory, reflecting real clinical complexity. Although recent vision-language models (VLMs) such as GPT-4o~\cite{achiam2023gpt}, InternVL3~\cite{zhu2025internvl3}, and Qwen2.5-VL~\cite{bai2025qwen2} have advanced multimodal reasoning, most medical VQA systems still rely on supervised fine-tuning (SFT), which depends on costly expert annotations and often encourages shallow pattern imitation instead of genuine semantic understanding—limitations that are particularly critical in clinical settings.

Reinforcement learning (RL), particularly reinforcement fine-tuning (RFT) with rule-based reward functions, has emerged as a promising alternative to conventional supervised approaches~\cite{guo2025deepseek,team2025qwq,wang2025visualprm}. By explicitly guiding the optimization process through reward signals, RFT enhances the reasoning and generalization capabilities of vision-language models while simultaneously reducing reliance on expensive human supervision. As illustrated in Figure~\ref{fig:problem}(a) and Figure~\ref{fig:problem}(b) , open-ended medical VQA better reflects real-world clinical scenarios, where responses must be flexible, context-aware, and explanatory rather than restricted to fixed choices. However, vanilla RL frameworks are not inherently designed to handle open-ended responses, as they rely on discrete or well-defined reward structures. This limitation makes it challenging to design clinically meaningful and reliable reward functions that can accurately assess the quality of medical reasoning and expression.

Traditional scores like BLEU~\cite{papineni2002bleu} and ROUGE~\cite{lin2004rouge} depend on simple surface word overlap. In medical settings, even minor lexical differences can imply drastically different meanings, risking the reward of incorrect but superficially similar responses~\cite{jain2021radgraphextractingclinicalentities}. Semantic-based metrics like BERTScore~\cite{zhang2019bertscore} and cosine similarity better capture meaning~\cite{rui2025improving}, but when used statically, they suffer from reward collapse~\cite{song2023reward}, where semantically distinct answers receive similar scores. As shown in Figure~\ref{fig:problem}(c), this results in flattened reward distributions, weak gradients, and reduced optimization efficiency, hindering learning and leading to suboptimal optimization paths.


To address the issues above, we propose \textbf{A}daptive \textbf{R}einforcement for \textbf{Med}ical Reasoning (\textbf{ARMed}), a novel and comprehensive reinforcement learning framework that effectively enhances open-ended medical reasoning. 
ARMed first incorporates domain-specific diagnostic knowledge via chain-of-thought supervision~\cite{wei2022chain}, enabling the model to produce interpretable, step-by-step reasoning traces instead of shallow pattern matching. 
Building on this foundation, ARMed applies Group Relative Policy Optimization (GRPO)~\cite{shao2024deepseekmath}, guided by an adaptive and clinically semantic reward that jointly evaluates textual correctness and semantic alignment. 
Unlike static semantic metrics that often trigger reward collapse, our adaptive reward dynamically and robustly scales its intensity according to inter-sample variance, amplifying distinctions between clinically meaningful responses while suppressing noisy or redundant feedback (Figure~\ref{fig:problem}(d)). 
Furthermore, ARMed follows a three-stage adaptive reinforcement pipeline—reward-driven pretraining, knowledge-augmented fine-tuning, and reward-based refinement—to progressively improve both factual accuracy and reasoning robustness. 
This integrated framework paradigm not only stabilizes optimization and mitigates semantic reward collapse but also aligns model behavior with authentic clinical reasoning, yielding more consistent, generalizable, and reliable performance across diverse and clinically relevant medical VQA benchmarks.

In summary, our main contributions are as follows:
\begin{itemize}
  \item We identify and formalize reward collapse in static semantic reward schemes during reinforcement learning.
  \item We propose \textbf{ARMed}, a framework that improves semantic reward discriminability through adaptive scaling based on Group Relative Policy Optimization (GRPO).
  \item We perform comprehensive experiments on six benchmarks, demonstrating significant improvements in both accuracy and generalization.
\end{itemize}

\section{Related Work}
\begin{figure*}[t]
\centering
\includegraphics[width=\linewidth]{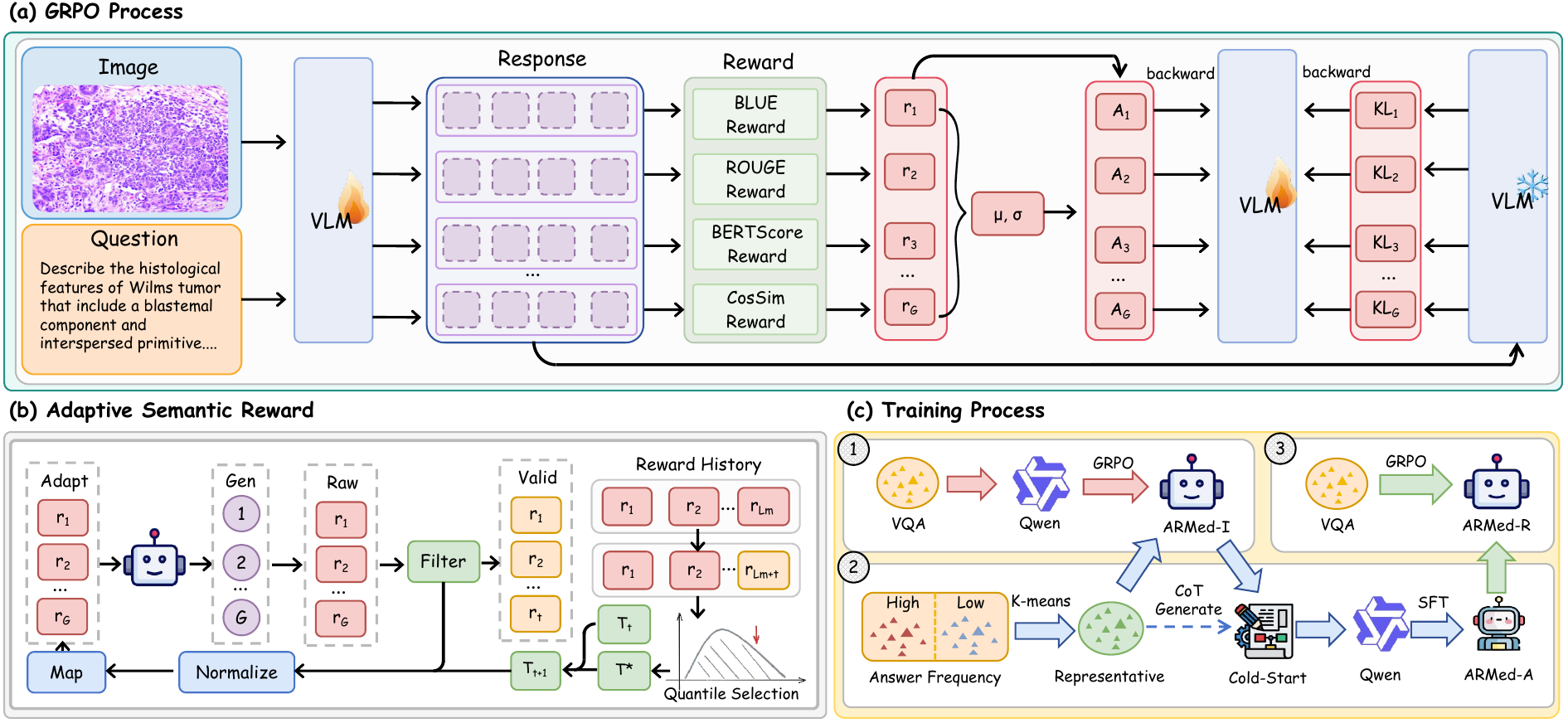} 
\caption{Overview of the ARMed framework in a medical QA task. (a) illustrates the GRPO training process; (b) shows how ARMed’s adaptive semantic reward mitigates reward collapse; and (c) outlines the three-stage training pipeline: Reward-driven Pretraining, Knowledge-enhanced Fine-tuning, and Reward-based Refinement.}

\label{fig:main}
\vspace{-0.5em}
\end{figure*}

\vspace{2pt}\noindent\textbf{Medical Vision-Language Models.}\vspace{1pt}
Recent progress in vision-language models (VLMs) has sparked growing interest in medical adaptation and multimodal clinical reasoning~\cite{chen2025forging,han2025beyond}. Models such as LLaVA-Med~\cite{li2023llava} and HuatuoGPT-Vision~\cite{chen2024huatuogpt} demonstrate strong performance in radiology-focused VQA and diagnostic reasoning tasks. However, most existing medical VLMs still rely on supervised fine-tuning (SFT) with final-answer annotations~\cite{Zhang_2024,liu2025application}. This paradigm is inherently rigid and resource-intensive: it requires massive, privacy-sensitive datasets, provides no explicit supervision for intermediate reasoning, and often leads to overfitting and weak generalization to out-of-distribution (OOD) data~\cite{wang2025large}. These limitations underscore the urgent need for more flexible, robust, and interpretable optimization frameworks beyond SFT.

\vspace{2pt}\noindent\textbf{Reinforcement Learning for General LLMs and VLMs.}\vspace{1pt}
Reinforcement learning (RL) has shown great potential in enhancing reasoning, alignment, and controllability for large-scale models. 
While PPO-based RLHF~\cite{schulman2017proximal} aligns models with human preferences, it depends on costly preference data and suffers from unstable critic estimation. 
Group Relative Policy Optimization (GRPO)~\cite{shao2024deepseekmath} effectively addresses these issues through critic-free, comparison-based updates, and DAPO~\cite{yu2025dapo} further improves overall training stability via dynamic sampling and token-level reward shaping. Taken together, these recent advances suggest that comparison-based RL can achieve more robust and more efficient optimization even under weak supervision.

\vspace{2pt}\noindent\textbf{Reinforcement Learning in the Medical Domain.}\vspace{1pt}
Reinforcement learning (RL) has recently been introduced to enhance clinical reasoning, factual consistency, and overall robustness in medical LLMs and VLMs~\cite{deng2025boostinggeneralizationreasoningvision,tan2025reasonrftreinforcementfinetuningvisual,ma2025rethinkingrlscalingvision}.
MedReason~\cite{wu2025medreasonelicitingfactualmedical} and HuatuoGPT-o1~\cite{chen2024huatuogpto1} improve reasoning chains through structured and verifiable supervision.
However, most existing medical VLMs rely on limited or static reward designs.
For instance, Med-R1~\cite{lai2025med} and MedVLM-R1~\cite{pan2025medvlm} restrict learning to multiple-choice visual QA, lacking open-ended reasoning capabilities.
GEMeX-ThinkVG~\cite{liu2025gemex} employs large medical LLMs as reward evaluators, leading to high computational cost and poor interpretability.
MedCCO~\cite{rui2025improving} adopts a static semantic similarity metric as the reward, but ignores the issue of reward collapse caused by insufficient semantic discrimination.
Consequently, these methods often fail to capture fine-grained semantic differences between responses, thereby limiting their ability to guide medical reasoning effectively.

To overcome these limitations, we adopt a GRPO-based framework with adaptive semantic rewards that explicitly emphasize inter-sample differences and contextual relevance.
This design substantially enhances reward discriminability and effectively mitigates reward collapse, thereby enabling more semantically grounded, reliable, and generalizable medical reasoning in VLMs.

\section{Methodology}

\subsection{Overview}
In this work, we present \textbf{ARMed}, an extension of the GRPO framework specifically tailored to enhance the reasoning and interpretability of models in open-ended medical visual question answering.
As shown in Figure~\ref{fig:main}(c), ARMed employs a three-stage training paradigm:
(1) \textbf{Reward-driven Pretraining}, where the base model is trained with a designed reward function for open-ended QA to produce the foundational reasoning model ARMed-Init (ARMed-I);
(2) \textbf{Knowledge-enhanced Fine-tuning}, in which ARMed-I generates explicit reasoning chains for knowledge-intensive medical samples, incorporated into a knowledge-augmented dataset used for supervised fine-tuning to obtain the knowledge-injected model ARMed-Augment (ARMed-A); and
(3) \textbf{Reward-based Refinement}, where ARMed-A undergoes further reward-driven optimization, yielding the expert-level model ARMed-Reasoner (ARMed-R).
By integrating reward-based training with explicit knowledge augmentation, ARMed significantly improves both the reasoning depth and factual reliability of models on open-ended medical VQA tasks.

\subsection{Preliminaries}

Group Relative Policy Optimization (GRPO) is a reinforcement learning algorithm designed to enhance the model's reasoning ability, as shown in Figure~\ref{fig:main}(a). Compared with the traditional proximal policy optimization (PPO) method, GRPO has significant differences in the following two aspects: First, GRPO implements a value-neutral training mechanism by introducing a generalized advantage estimation (GAE) based on group relative rewards; second, its reward signal comes from a verifiable rule-based evaluation criterion rather than relying on a pre-trained reward model, thereby improving the interpretability and reliability of the reward. This also allows GPRO to improve computational efficiency while reducing computing resource usage~\cite{shao2024deepseekmath}.

During each optimization step, the policy $\pi_{\theta_{\text{old}}}$ produces a batch of $G$ output candidates $\{o_i\}_{i=1}^G$. Each output candidate receives a numerical reward $r_i$ calculated according to predefined scoring rules. The advantage $A_i$ is computed by standardizing the rewards within the group, achieved by subtracting the group's average reward and dividing by the corresponding standard deviation:
\begin{equation}
\small
A_i = \frac{r_i - \text{mean}(\{r_j\}_{j=1}^G)}{\text{std}(\{r_j\}_{j=1}^G)}.
\label{eq:advantage}
\end{equation}

Formally, denote $P(Q)$ as the distribution of questions used during training, with $q$ representing a sampled query at the current iteration. Let $\pi_{\theta_{\text{old}}}$ and $\pi_{\theta_{\text{new}}}$ refer to the previous and updated policies, respectively, and let $o$ denote a complete response drawn from the policy. Additionally, define $\pi_{\theta_{\text{ref}}}$ as a fixed reference policy, which corresponds to the frozen base VLM. Suppose $G$ represents the number of responses sampled per question in each iteration. The objective function for GRPO is then expressed as:

\begin{equation}
\small
\begin{split}
\mathcal{J}_{\text{GRPO}}(\theta) =
&\; \mathbb{E}_{q \sim P(Q),\, \{o_l\}_{l=1}^G \sim \pi_{\theta_{\text{old}}}(O|q)} \\
&\frac{1}{G} \sum_{i=1}^G \frac{1}{|o_i|} \sum_{t=1}^{|o_i|} \Bigg[ 
\min \Bigg( 
\frac{\pi_{\theta}(o_{i,t}|q_i, o_{i,<t})}
{\pi_{\theta_{\text{old}}}(o_{i,t}|q_i, o_{i,<t})} \hat{A}_{i,t}, \\
&\text{clip} \Bigg(
\frac{\pi_{\theta}(o_{i,t}|q_i, o_{i,<t})}
{\pi_{\theta_{\text{old}}}(o_{i,t}|q_i, o_{i,<t})},\, 
1-\epsilon,\, 1+\epsilon \Bigg) \hat{A}_{i,t} 
\Bigg) \\
& - \beta\, D_{\mathrm{KL}}\big[\pi_{\theta} \,\|\, \pi_{\text{ref}}\big] \Bigg],
\end{split}
\end{equation}
where $\mathbb{D}_{\text{KL}}(\pi_{\theta} \| \pi_{\text{ref}})$ serves as a regularization term to penalize divergence from the reference policy $\pi_{\text{ref}}$, with $\beta$ controlling its strength.

\subsection{Adaptive Reward Function Design}

\label{Adaptive Reward Function Design}

For open-ended medical question answering, we design a reward function consisting of textual correctness reward, adaptive semantic alignment reward, and format reward.

\begin{figure}[htbp]
\centering
\includegraphics[width=\linewidth]{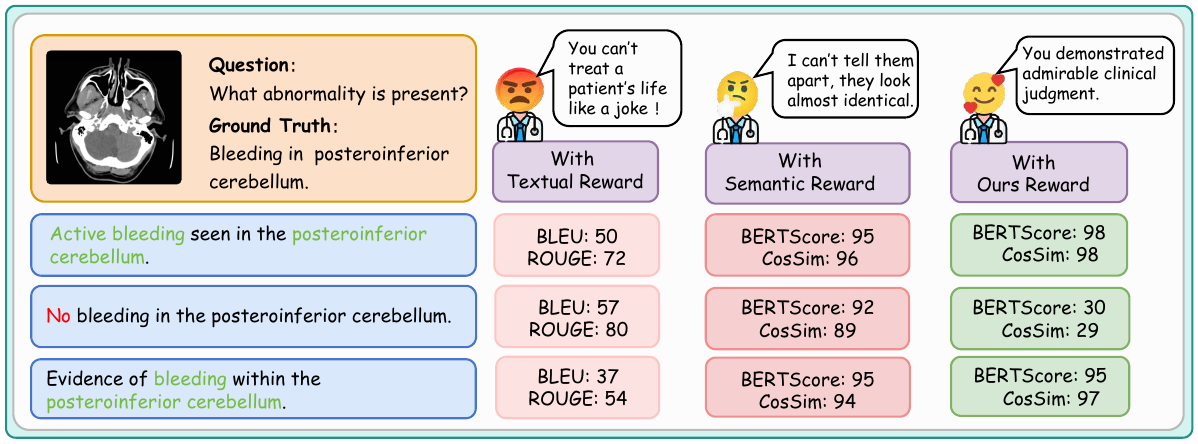} 
\caption{Example of ARMed in a medical QA task. This illustrates why textual similarity alone is insufficient, how naive semantic rewards can collapse, and how ARMed enables stable and meaningful semantic evaluation for reliable medical reasoning.}
\label{fig:problemdemo}
\vspace{-0.5em}
\end{figure}
As shown in Figure~\ref{fig:problemdemo}, the example demonstrates how relying solely on textual similarity can fail to capture the true semantic alignment in medical reasoning. It highlights that naive semantic rewards often collapse when faced with subtle yet clinically significant meaning differences, resulting in unstable or misleading optimization signals. This motivates the need for our adaptive semantic reward, which preserves meaningful distinctions and ensures stable guidance for policy improvement.

From a theoretical perspective, the effectiveness of our adaptive semantic reward lies in its role within the composite reward used by GRPO. Although the advantage $A_i$ in GRPO is computed by normalizing rewards within a sampled group (\textit{i.e. }Equation~\ref{eq:advantage}), the key limitation arises when the semantic reward itself lacks discriminability. In such cases, its signal becomes indistinguishable from other reward components and is effectively overshadowed by those with stronger variance, contributing little to the overall policy update. Our adaptive reward mitigates this issue by dynamically reshaping the semantic reward distribution to enhance inter-sample distinctions, thereby amplifying its relative contribution to the total reward signal. This adjustment improves gradient informativeness while preserving the original reward ranking, thus remaining consistent with the policy invariance principle~\cite{ng1999policy}.

\vspace{2pt}\noindent\textbf{Textual Correctness Reward}.\vspace{1pt}
Traditional accuracy and exact match metrics are strict matching methods that reward the generated text only when it exactly matches the reference answer; otherwise, the reward is zero. While this approach can provide effective supervision for closed-ended question answering, it often leads to sparse and unstable rewards in open-ended settings due to its stringent criteria.
To mitigate this issue, especially during early reinforcement learning when model predictions deviate considerably from references, we adopt BLEU-1~\cite{papineni2002bleu} and ROUGE-1~\cite{lin2004rouge}. These n-gram overlap metrics provide graded and dense feedback, ensuring non-zero rewards even for partially correct generations. This alleviates early-stage reward sparsity and facilitates smoother policy optimization while maintaining sensitivity to textual correctness.
\small
\begin{equation}
\begin{split}
     R_c(o, g) =& \lambda_1 \cdot \text{BLEU}_1(o, g) \\
     +& (1 - \lambda_1) \cdot \text{ROUGE}_1(o, g),
\end{split}  
\end{equation}
where $o$ denotes the predicted answer, $g$ is the ground truth answer, and $\lambda_1 \in [0,1]$ is a weighting factor balancing the contributions of BLEU-1 and ROUGE-1 scores.

\vspace{2pt}\noindent\textbf{Adaptive Semantic Alignment Reward}.\vspace{1pt} A key challenge in medical open-ended QA lies in the diversity of expressions. 
Semantically equivalent responses may differ in wording or syntax, yet overlap metrics (\textit{e.g.}, ROUGE, BLEU) fail to capture such equivalence and thus unfairly penalize valid variations.

To address this, we define a semantic alignment reward that precisely measures the semantic similarity between the generated answer $o$ and reference $g$:

\small
\begin{equation}
\begin{split}
R_s(o,g) &= \lambda_2 \cdot \text{BERTScore}(o,g) \\
    &+ (1-\lambda_2) \cdot \text{CosSim}(o,g),
\end{split}
\end{equation}

\vspace{2pt}\noindent where $\lambda_2 \in [0,1]$ balances token-level contextual similarity (BERTScore) and sentence-level embedding similarity (CosSim). 

However, directly using $R_s$ is problematic: (1) scores often cluster within narrow, high-value ranges, making it difficult to distinguish response quality; and (2) reward scales vary across tasks and models, lacking consistent thresholds. Thus, absolute rewards are unstable and weakly discriminative. To address this, we propose an adaptive reward calibration that combines dynamic historical statistics with nonlinear mapping (Fig.~\ref{fig:main}(a)). Given the historical reward buffer $\mathcal{H}_t$ and current batch ${r_i}$, we retain values above a fraction $\rho T_t$ of the current threshold:
\begin{equation}
\mathcal{R}_t = \{ r_i \mid r_i > \rho T_t \}.
\end{equation}
The buffer is updated (truncated to $L_{\max}$), and a target threshold $T^*$ is set as the $p$-th percentile of $\mathcal{H}_t$. The threshold then smoothly adapts with step limit $\delta_{\max}$:
\small
\begin{equation}
\begin{split}
T_{t+1} = \text{clip}\!\big(&\,T_t + 
    \text{clip}(T^*-T_t,\,-\delta_{\max},\,\delta_{\max}),\\
    & T_{\min},\, T_{\max}\big).
\end{split}
\end{equation}

\vspace{2pt}\noindent Each reward is normalized relative to the updated threshold:
\small
\begin{equation}
\hat{r}_i = \text{clip}\!\left(\frac{r_i - T_{t+1}}{1 - T_{t+1} + \epsilon}, -1, 1\right),
\end{equation}
\vspace{2pt}\noindent and passed through an asymmetric S-shaped mapping to enhance sensitivity near the threshold:
\small
\begin{equation}
r_i' = 0.5 \big[1 + \tanh(\alpha_s \hat{r}_i)\big], \quad
\alpha_s =
\begin{cases}
\alpha_{\text{pos}}, & \hat{r}_i \ge 0,\\
\alpha_{\text{neg}}, & \hat{r}_i < 0.
\end{cases}
\end{equation}

\vspace{2pt}\noindent Denoting the above transformation as $\text{Adapt}(\cdot)$, the final adaptive semantic alignment reward is given by:
\begin{equation}
\begin{split}
R_{as}(o,g) =&\, \lambda_2 \cdot \text{Adapt}(\text{BERTScore}(o,g))\\
             &+ (1-\lambda_2) \cdot \text{Adapt}(\text{CosSim}(o,g)).
\end{split}
\end{equation}

\vspace{2pt}\noindent This adaptive design normalizes reward ranges, improves discrimination near semantic boundaries, and ensures stable alignment across tasks and models.

\vspace{2pt}\noindent\textbf{Format Reward}.\vspace{1pt}To ensure a well-structured output, we impose a format reward, denoted as \( R_f \), which verifies whether the generated content conforms to the required tagging scheme. Specifically, the reasoning process must be enclosed within the \texttt{<think> \ldots </think>} tags, while the final answer should be contained within the \texttt{<answer> \ldots </answer>} tags.

\vspace{2pt}\noindent\textbf{Total Reward}.\vspace{1pt}Total reward can be computed as:
\begin{equation}
R_{\text{total}}=\frac{\gamma_1\cdot R_c+ \gamma_2\cdot R_{as}+ \gamma_3 \cdot R_f}{\gamma_1+\gamma_2+\gamma_3},
\end{equation}
where \(\gamma_1, \gamma_2,\) and \(\gamma_3\) are non-negative weighting coefficients that determine the relative importance of each reward component.

\begin{table*}[!t]
\caption{Performance comparison among vision–language models on in-domain medical VQA benchmarks. For each metric, \textbf{bold} and \underline{underlined} entries indicate the best and second-best results, respectively. \setlength{\fboxsep}{2pt}\colorbox{gray!20}{Gray-shaded} columns correspond to dataset-specific averages, while \setlength{\fboxsep}{2pt}\colorbox{blue!15}{blue-shaded} columns correspond to averages across all metrics.}
\centering
\setlength{\tabcolsep}{3pt}
\resizebox{\textwidth}{!}{%
\begin{tabular}{@{}l*{5}{c}*{5}{c}*{5}{c}c@{\hspace{\tabcolsep}}}
\toprule
\multirow{2}{*}{Model} & \multicolumn{5}{c}{Path-VQA} & \multicolumn{5}{c}{SLAKE} & \multicolumn{5}{c}{VQA-RAD} & \multirow{2}{*}{\textbf{Avg.}} \\ 
\cmidrule(lr){2-6} \cmidrule(lr){7-11} \cmidrule(lr){12-16}
& BLEU-1 & ROUGE-1 & BERTScore & CosSim & Avg. & BLEU-1 & ROUGE-1 & BERTScore & CosSim & Avg. & BLEU-1 & ROUGE-1 & BERTScore & CosSim & Avg. & \\
\midrule
\multicolumn{17}{c}{\textit{General VLM}} \\ 
\midrule
Qwen2.5-VL-7B & 5.25 & 8.24 & 88.45 & 85.03 & \cellcolor{gray!20}46.74 & 10.50 & 15.24 & 91.73 & 90.13 & \cellcolor{gray!20}51.90 & 10.17 & 14.77 & 91.18 & 89.03 & \cellcolor{gray!20}51.29 & \cellcolor{blue!15}49.98 \\
LLaVA-v1.6-7B & 19.27 & 20.86 & 87.12 & 83.91 & \cellcolor{gray!20}52.79 & 26.30 & 28.99 & 91.41 & 89.78 & \cellcolor{gray!20}59.12 & 26.10 & 28.12 & 90.14 & 88.30 & \cellcolor{gray!20}58.17 & \cellcolor{blue!15}56.69 \\
LLaVA-v1.6-13B & 7.23 & 10.04 & 85.55 & 81.66 & \cellcolor{gray!20}46.12 & 18.49 & 22.57 & 90.82 & 88.86 & \cellcolor{gray!20}55.19 & 14.65 & 17.73 & 89.23 & 87.04 & \cellcolor{gray!20}52.16 & \cellcolor{blue!15}51.16 \\
InternVL3-2B & 28.56 & 30.38 & 92.71 & 88.51 & \cellcolor{gray!20}60.04 & 39.18 & 43.49 & 93.59 & 91.67 & \cellcolor{gray!20}66.98 & 35.92 & 38.72 & 92.17 & 90.01 & \cellcolor{gray!20}64.21 & \cellcolor{blue!15}63.74 \\
InternVL3-8B & 4.53 & 6.78 & 80.66 & 78.29 & \cellcolor{gray!20}42.57 & 17.36 & 23.62 & 87.29 & 85.38 & \cellcolor{gray!20}53.41 & 10.42 & 14.25 & 84.50 & 81.82 & \cellcolor{gray!20}47.75 & \cellcolor{blue!15}47.91 \\
InternVL3-14B & 3.70 & 6.08 & 80.37 & 77.80 & \cellcolor{gray!20}41.99 & 16.87 & 22.32 & 88.62 & 87.02 & \cellcolor{gray!20}53.71 & 10.86 & 16.18 & 87.40 & 85.81 & \cellcolor{gray!20}37.62 & \cellcolor{blue!15}44.44 \\ 
\midrule
\multicolumn{17}{c}{\textit{Medical VLM}} \\ 
\midrule
LLaVA-Med-7B & 2.39 & 4.50 & 85.66 & 84.15 & \cellcolor{gray!20}44.18 & 3.29 & 7.20 & 89.25 & 88.29 & \cellcolor{gray!20}47.01 & 3.18 & 6.07 & 89.09 & 87.95 & \cellcolor{gray!20}46.57 & \cellcolor{blue!15}45.92 \\
HuatuoGPT-V-7B & 6.59 & 10.12 & 83.10 & 80.22 & \cellcolor{gray!20}45.01 & 13.11 & 20.25 & 88.23 & 86.26 & \cellcolor{gray!20}51.96 & 7.16 & 11.62 & 83.75 & 80.90 & \cellcolor{gray!20}45.86 & \cellcolor{blue!15}47.61 \\ 
\midrule
\multicolumn{17}{c}{\textit{Fine-tuned VLM}} \\ 
\midrule
Qwen2.5-VL-3B & 5.30 & 8.20 & 86.91 & 85.10 & \cellcolor{gray!20}46.38 & 13.34 & 19.09 & 90.35 & 89.14 & \cellcolor{gray!20}52.98 & 9.58 & 13.38 & 88.46 & 87.64 & \cellcolor{gray!20}49.77 & \cellcolor{blue!15}49.71 \\
\textit{w}. SFT & \underline{55.43} & \underline{56.62} & 97.74 & 94.36 & \cellcolor{gray!20}\underline{76.04} & \underline{70.58} & \underline{71.54} & 99.10 & 97.67 & \cellcolor{gray!20}\underline{84.72} & \underline{56.58} & \underline{58.68} & \underline{98.36} & 96.27 & \cellcolor{gray!20}\underline{77.47} & \cellcolor{blue!15}\underline{79.41} \\
\textit{w}. GRPO & 52.36 & 54.28 & \underline{98.04} & \underline{95.60} & \cellcolor{gray!20}75.07 & 68.30 & 69.49 & \underline{99.13} & \underline{97.89} & \cellcolor{gray!20}83.70 & 51.43 & 53.68 & 97.54 & \underline{97.28} & \cellcolor{gray!20}74.98 & \cellcolor{blue!15}77.92 \\
ARMed-R(Ours) & \textbf{63.61} & \textbf{64.96} & \textbf{98.46} & \textbf{95.81} & \cellcolor{gray!20}\textbf{80.71} & \textbf{76.14} & \textbf{76.65} & \textbf{99.46} & \textbf{98.13} & \cellcolor{gray!20}\textbf{87.60} & \textbf{58.72} & \textbf{61.10} & \textbf{98.69} & \textbf{97.29} & \cellcolor{gray!20}\textbf{78.95} & \cellcolor{blue!15}\textbf{82.42} \\ 
\bottomrule
\end{tabular}%
}

\label{tab:all_indomain_performance}

\vspace{-0.5em}
\end{table*}

\subsection{Medical Thinking Knowledge Injection}

\noindent\textbf{Medical Thinking Knowledge Injection.}\vspace{1pt}
During reinforcement learning for open-ended question answering, we observe a critical \textit{reward-driven bias}: the model tends to overfit to answer types previously associated with high rewards, even when incorrect. 
Formally, given a question–answer pair $(x_i, y_i)$ and its reward $r_i$, the expected policy gradient is dominated by high-reward samples:
\begin{equation}
\nabla_\theta J(\theta) \propto \sum_i r_i \nabla_\theta \log \pi_\theta(y_i|x_i),
\label{eq:reward-bias}
\end{equation}
which causes the model to disproportionately reinforce rewarded answers. 
When encountering semantically similar questions, it often reproduces these biased high-reward responses while ignoring correct alternatives with low historical rewards. 
In medical domains, such behavior is hazardous, as erroneous responses may distort clinical reasoning and jeopardize patient safety.

\smallskip
\noindent\textbf{Knowledge Construction.}\vspace{1pt}
To mitigate this bias, we propose a medical thinking knowledge injection mechanism that enhances knowledge diversity and reasoning robustness. 
Let $\mathcal{D} = \{(x_i, y_i)\}$ denote the QA corpus, and let $f(y_i)$ represent the frequency of each unique answer. 
We define the \textit{core knowledge set} as:
\begin{equation}
\mathcal{D}_{\text{core}} = \{(x_i, y_i) \mid f(y_i) > \tau\},
\label{eq:core}
\end{equation}
where $\tau$ is a frequency threshold. 
The remaining QA pairs,
\begin{equation}
\mathcal{D}_{\text{sup}} = \{(x_i, y_i) \mid f(y_i) \le \tau\},
\label{eq:sup}
\end{equation}
are retained as \textit{supplementary knowledge} to capture long-tail clinical reasoning patterns. 
Each QA pair serves as a semantic anchor, where answers guide the contextual reasoning process.

To ensure diversity, we embed questions using a Sentence Transformer $E(x_i)$ and apply K-Means clustering within each high-frequency group:
\begin{equation}
\mathcal{C} = \text{KMeans}\big(\{E(x_i) \mid (x_i, y_i) \in \mathcal{D}_{\text{core}}\}\big),
\end{equation}
then select, from each cluster $c_k \in \mathcal{C}$, the representative QA pair whose question embedding is closest to the centroid $\mu_k$:
\begin{equation}
(x_k^*, y_k^*) = \arg\min_{(x_i, y_i) \in c_k} \|E(x_i) - \mu_k\|_2.
\label{eq:representative}
\end{equation}
This yields a compact yet diverse \textit{medical knowledge base} for reasoning injection.

\smallskip
\noindent\textbf{Knowledge Injection and Training.}\vspace{1pt}
We leverage the baseline reasoning model trained with our adaptive reward, denoted as \textbf{ARMed-Init (ARMed-I)}, to expand chain-of-thought reasoning for $\mathcal{D}_{\text{core}}$. 
The resulting augmented dataset $\mathcal{D}_{\text{aug}} = \{(x_i, \hat{y}_i)\}$ contains richer, stepwise reasoning traces. 
Unlike mathematical reasoning datasets that mainly encode structural logic, $\mathcal{D}_{\text{aug}}$ emphasizes \textit{domain-specific clinical knowledge} and linguistic conventions. 
We fine-tune the model on $\mathcal{D}_{\text{aug}}$ to obtain \textbf{ARMed-Augment (ARMed-A)}, and further apply the same reinforcement learning procedure as in ARMed-I to produce the final \textbf{ARMed-Reasoner (ARMed-R)} — capable of robust, knowledge-grounded medical reasoning.

\begin{table}[!htbp]
\caption{Performance comparison among vision–language models on out-of-domain medical VQA benchmarks. For each metric, \textbf{bold} and \underline{underlined} entries indicate the best and second-best results, respectively. \setlength{\fboxsep}{2pt}\colorbox{gray!20}{Gray-shaded} columns correspond to dataset-specific averages, while \setlength{\fboxsep}{2pt}\colorbox{blue!15}{blue-shaded} columns correspond to averages across all metrics.}
\centering
\setlength{\tabcolsep}{3pt}
\resizebox{\linewidth}{!}{%
\begin{tabular}{@{}l*{5}{c}cc c@{\hspace{\tabcolsep}}}
\toprule
\multirow{2}{*}{Model} & \multicolumn{5}{c}{VQA-Med} & PMC-VQA  & MedXpert & \multirow{2}{*}{\textbf{Avg.}} \\ 
\cmidrule(lr){2-6}
 & {\small BLEU-1} & {\small ROUGE-1} &{\small BERTScore}  & {\small CosSim} & Avg. & (Acc.) & (Acc.) &  \\
\midrule
\multicolumn{9}{c}{\textit{General VLM}} \\
\midrule
Qwen2.5-VL-7B & 6.64 & 10.40 & 89.53 & 87.36 & \cellcolor{gray!20}48.48 & 47.80 & 15.60 & \cellcolor{blue!15}37.29 \\
LLaVA-v1.6-7B & 3.76 & 5.49 & 84.41 & 80.22 & \cellcolor{gray!20}43.47 & 33.05 & 11.25 & \cellcolor{blue!15}29.26 \\
LLaVA-v1.6-13B & 3.76 & 5.47 & 86.08 & 81.41 & \cellcolor{gray!20}44.18 & 34.30 & 13.80 & \cellcolor{blue!15}30.76 \\
InternVL3-2B & 21.44 & 22.82 & 94.93 & 90.43 & \cellcolor{gray!20}57.41 & 40.69 & 18.95 & \cellcolor{blue!15}39.02 \\
InternVL3-8B & 14.80 & 18.16 & 85.79 & 82.19 & \cellcolor{gray!20}50.24 & 48.55 & 21.09 & \cellcolor{blue!15}39.96 \\
InternVL3-14B & 10.69 & 14.99 & 86.18 & 82.92 & \cellcolor{gray!20}48.70 & \underline{48.75} & 15.05 & \cellcolor{blue!15}37.50 \\
\midrule
\multicolumn{9}{c}{\textit{Medical VLM}} \\
\midrule
LLaVA-Med-7B & 2.04 & 3.93 & 86.63 & 84.31 & \cellcolor{gray!20}44.23 & 23.80 & 20.05 & \cellcolor{blue!15}29.36 \\
HuatuoGPT-V-7B & 5.70 & 10.20 & 82.33 & 80.73 & \cellcolor{gray!20}44.74 & \textbf{50.00} & 21.65 & \cellcolor{blue!15}38.80 \\
\midrule
\multicolumn{9}{c}{\textit{Fine-tuned VLM}} \\
\midrule
Qwen2.5-VL-3B & 4.88 & 6.39 & 69.47 & 67.01 & \cellcolor{gray!20}36.94 & 45.70 & 11.60 & \cellcolor{blue!15}31.41 \\
\textit{w}. SFT & \textbf{22.44} & \textbf{24.55} & 95.54 & 90.11 & \cellcolor{gray!20}\underline{58.16} & 46.80 & 18.35 & \cellcolor{blue!15}\underline{41.10} \\
\textit{w}. GRPO & 19.95 & 20.85 & \underline{96.32} & \underline{91.71} & \cellcolor{gray!20}57.21 & 48.10 & \underline{20.65} & \cellcolor{blue!15}41.99 \\
ARMed-R(Ours) & \underline{21.51} & \underline{23.17} & \textbf{96.56} & \textbf{92.36} & \cellcolor{gray!20}\textbf{58.40} & \underline{48.75} & \textbf{22.30} & {\cellcolor{blue!15}\textbf{43.15}} \\
\bottomrule
\end{tabular}%
}

\label{tab:all_outdomain_performance}
\end{table}

\section{Experiments}

\subsection{Datasets}
We use  VQA-RAD~\cite{lau2018dataset}, SLAKE~\cite{liu2021slake}, and PathVQA~\cite{he2020pathvqa} as training datasets and in-domain test datasets, and VQA-Med~\cite{ben2019vqa}, PMC-VQA~\cite{zhang2023pmc} and MedXpertQA~\cite{zuo2025medxpertqa} as out-of-domain test datasets. Following \citeauthor{rui2025improving}, we perform VQA redefinition on the in-domain training and test sets. Please see the Appendix for a detailed introduction to the dataset.

\begin{figure}[htbp]
    \centering
    \includegraphics[width=\linewidth]{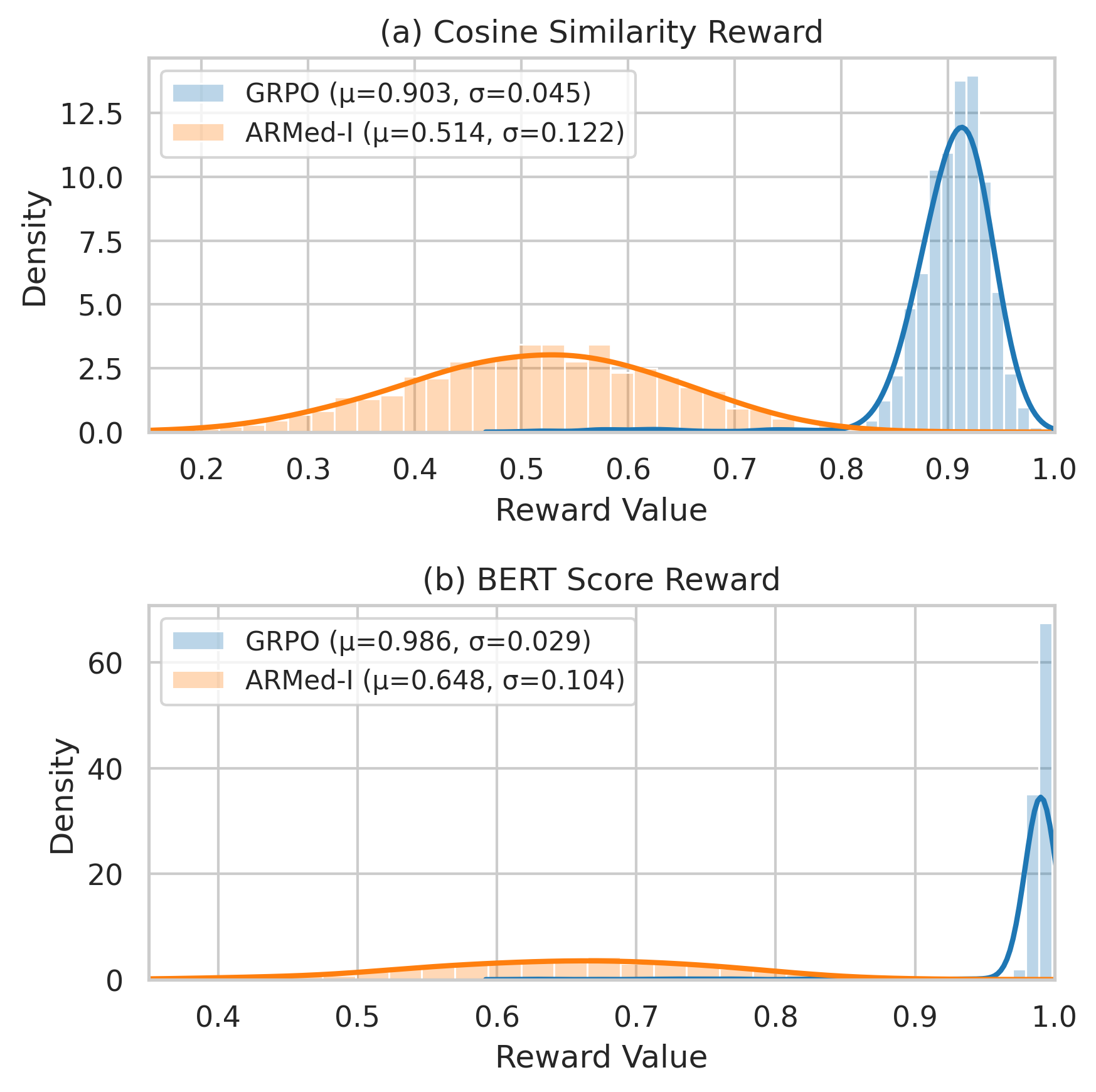}
    \caption{Distribution of model semantic rewards during training. It compares the distributions of non-discriminative (GRPO) and adaptive (ARMed-I) rewards.}
    \label{fig:Reward_Collapse}
    \vspace{-0.5em}
\end{figure}

\begin{table*}[!htbp]
\caption{Ablation study results on a subset of the test sets. 
\textbf{Aug} denotes \textit{augmentation}, 
\textbf{Text} denotes \textit{text reward}, 
\textbf{Sem} denotes \textit{semantic reward}, 
and \textbf{Ada} denotes \textit{adaptive semantic reward}. 
Complete results on all test sets can be found in the Appendix. 
For each metric column, \textbf{bold} denotes the best performance and \underline{underline} denotes the second-best. \setlength{\fboxsep}{2pt}\colorbox{gray!20}{Gray-shaded} columns correspond to dataset-specific averages, while \setlength{\fboxsep}{2pt}\colorbox{blue!15}{blue-shaded} columns correspond to averages across all metrics.}
\setlength{\tabcolsep}{3pt}
\centering
\resizebox{\linewidth}{!}{
\begin{tabular}{@{}l*{4}{c}*{5}{c}*{5}{c}cc c@{\hspace{\tabcolsep}}}
\toprule
\multirow{3}{*}{Model} & \multicolumn{4}{c}{Method} & \multicolumn{5}{c}{Path-VQA} & \multicolumn{5}{c}{SLAKE} & PMC VQA & MedXpert & \multirow{2}{*}{\textbf{Avg.}} \\
\cmidrule(lr){2-5}\cmidrule(lr){6-10}\cmidrule(lr){11-15}
 & \textit{w}.\ Aug & \textit{w}.\ Text & \textit{w}.\ Sem  & \textit{w}.\ Ada & BLEU-1 & ROUGE-1 & BERTScore & CosSim & Avg. & BLEU-1 & ROUGE-1 & BERTScore & CosSim & Avg. & (Acc.) & (Acc.) &  \\
\midrule
\multirow{8}{*}{Qwen2.5-VL-3B}
& \xmark & \xmark & \xmark & \xmark & 5.30 & 8.20 & 86.91 & 85.10 & \cellcolor{gray!20}46.38 & 13.34 & 19.09 & 90.35 & 89.14 & \cellcolor{gray!20}52.98 & 45.70 & 11.60 & \cellcolor{blue!15}46.07 \\
& \xmark & \cmark & \xmark & \xmark & 49.62 & 56.07 & 96.21 & 93.80 & \cellcolor{gray!20}73.93 & 64.91 & 68.76 & 98.48 & 97.19 & \cellcolor{gray!20}82.34 & 46.40 & 19.90 & \cellcolor{blue!15}68.99 \\
& \xmark & \cmark & \cmark & \xmark & 52.36 & 54.28 & \underline{98.04} & 95.60 & \cellcolor{gray!20}75.07 & 68.30 & 69.49 & 99.13 & 97.89 & \cellcolor{gray!20}83.70 & \underline{48.10} & 20.65 & \cellcolor{blue!15}70.29 \\
& \xmark & \cmark & \xmark & \cmark & 54.88 & 58.77 & 97.49 & 94.74 & \cellcolor{gray!20}76.47 & 68.44 & 70.49 & 98.91 & 97.54 & \cellcolor{gray!20}83.85 & 47.30 & 20.90 & \cellcolor{blue!15}70.49 \\
& \cmark & \xmark & \xmark & \xmark & 40.72 & 43.06 & 96.49 & 94.11 & \cellcolor{gray!20}68.60 & 59.39 & 60.54 & 97.76 & 97.32 & \cellcolor{gray!20}78.75 & 43.40 & 19.25 & \cellcolor{blue!15}65.63 \\
& \cmark & \cmark & \xmark & \xmark & \underline{61.72} & 61.92 & 97.65 & 95.74 & \cellcolor{gray!20}79.26 & 73.16 & \underline{74.87} & 98.85 & 97.15 & \cellcolor{gray!20}86.01 & 46.60 & 20.50 & \cellcolor{blue!15}72.05 \\
& \cmark & \cmark & \cmark & \xmark & 61.23 & \underline{62.53} & 97.64 & \textbf{96.03} & \cellcolor{gray!20}\underline{79.36} & \underline{73.22} & 73.96 & \underline{99.15} & \underline{97.97} & \cellcolor{gray!20}\underline{86.08} & 48.05 & \underline{20.95} & \cellcolor{blue!15}\underline{72.34} \\
& \cmark & \cmark & \xmark & \cmark & \textbf{63.61} & \textbf{64.96} & \textbf{98.46} & \underline{95.81} & \cellcolor{gray!20}\textbf{80.71} & \textbf{76.14} & \textbf{76.65} & \textbf{99.46} & \textbf{98.13} & \cellcolor{gray!20}\textbf{87.60} & \textbf{48.75} & \textbf{22.30} & \cellcolor{blue!15}\textbf{73.72} \\
\bottomrule
\end{tabular}
}

\label{tab:qwen25_ablation}
\end{table*}

\subsection{Model Zoo and Evaluation Metric}
We compare the performance of our ARMed  against a range of vision-language models (VLMs), including both general-purpose and medical-specific models. The general VLMs include:
(1) the Qwen2.5-VL family~\cite{bai2025qwen2}, including Qwen2.5-VL-3B and Qwen2.5-VL-7B;
(2) the InternVL3 family~\cite{zhu2025internvl3}, including InternVL3-2B, InternVL3-8B, and InternVL3-14B;
(3) the LLaVA-v1.6 family~\cite{liu2024improved}, including LLaVA-v1.6-7B and LLaVA-v1.6-13B.
We also evaluate against medical-domain VLMs, including LLaVA-Med~\cite{li2023llava} and HuatuoGPT-V~\cite{chen2024huatuogpt}.

To ensure both accurate evaluation and reliable semantic rewards in reinforcement learning, we employ PubMedBERT~\cite{pubmedbert} to compute BERTScore and BioBERT-mnli-snli-scinli-scitail-mednli-stsb~\cite{deka2022evidence} to calculate cosine similarity, both of which are pretrained on biomedical corpora and thus better aligned with the domain characteristics of medical texts.

For multiple-choice questions, we use accuracy (ACC) as the evaluation metric. For open-ended questions, we employ BLEU-1, ROUGE-1, BERTScore, and Cosine Similarity (CosSim) as evaluation metrics to comprehensively assess answer quality.

\subsection{Implementation Details}
All experiments are trained on $4\times$ H100 GPUs (80GB) with FlashAttention-2. The model is initialized from Qwen2.5-VL-3B-Instruct~\cite{bai2025qwen2} and fully fine-tuned with a learning rate of $1\times10^{-6}$. We use a per-GPU batch size of 2 with 2-step gradient accumulation (effective batch size 4). Input images are capped at 401k pixels, and the maximum generation length is 1{,}024 tokens.

For GRPO, we use MS-SWIFT~\cite{zhao2024swiftascalablelightweightinfrastructure} for distributed training; each policy samples 8 candidate responses per example with temperature $\tau=0.7$. The reward weights are $\lambda_1=0.5$, $\lambda_2=0.2$, $\gamma_1=0.4$, $\gamma_2=0.4$, and $\gamma_3=0.2$. Each configuration is trained for one epoch. SFT experiments are conducted with LLaMA Factory~\cite{zheng2024llamafactory}. Additional training details are provided in the supplementary material.

\section{Results and Discussion}

\begin{figure}[htbp]
    \centering
    \includegraphics[width=\linewidth]{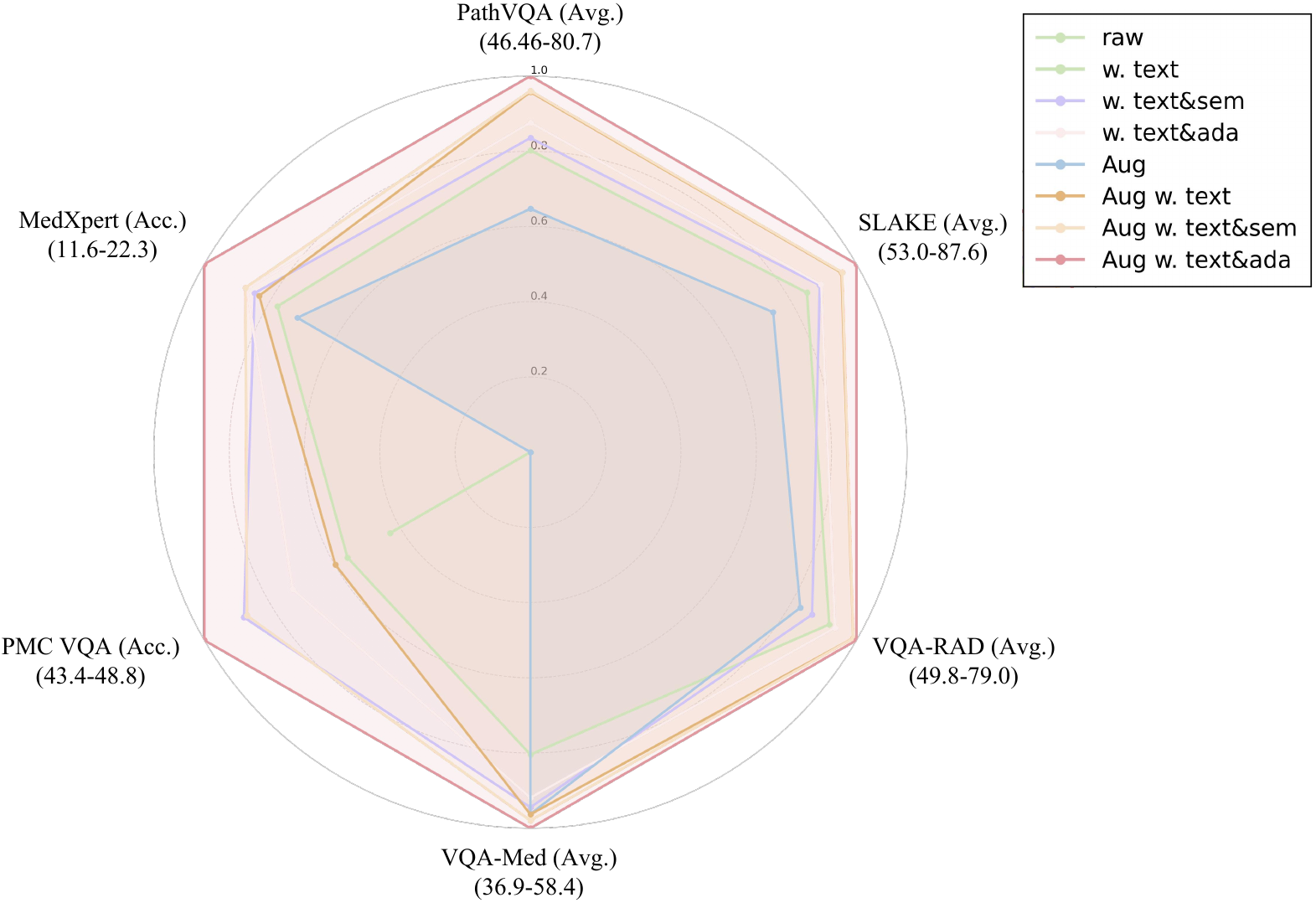}
    \caption{Ablation study results of different model variants on six medical benchmarks. “Avg.” denotes open-ended QA datasets (average of multiple metrics), and “acc.” represents closed-ended QA datasets (accuracy). The radar chart illustrates the contribution of text, semantic, and adaptive modules, as well as data augmentation strategies, to overall performance.}
    \label{fig:Model_Comparison}
     \vspace{-0.8em}
\end{figure}

\subsection{Overall Performance}

As shown in Table~\ref{tab:all_indomain_performance} and Table~\ref{tab:all_outdomain_performance}, \textbf{ARMed} achieves state-of-the-art results on both in-domain and out-of-domain medical VQA benchmarks, with average scores of \textbf{82.42\%} and \textbf{43.15\%}, respectively. It consistently outperforms both generalist and medical-specific baselines on five of six datasets, demonstrating strong reasoning ability and generalization across domains. On the in-domain test set, ARMed achieves a \textbf{20.67\%} improvement over the best-performing model InternVL3-2B, and a \textbf{3.19\%} gain over InternVL3-8B on the out-of-domain set. Notably, models such as InternVL3-14B and HuatuoGPT-Vision-7B that perform comparably possess more than twice the number of parameters, highlighting ARMed’s efficiency and scalability. These results confirm the robustness and effectiveness of our adaptive semantic reward–guided GRPO reinforcement learning, which enables more stable optimization and enhances clinically reliable reasoning.

\begin{figure*}[!htbp]
    \centering
    \includegraphics[width=\linewidth]{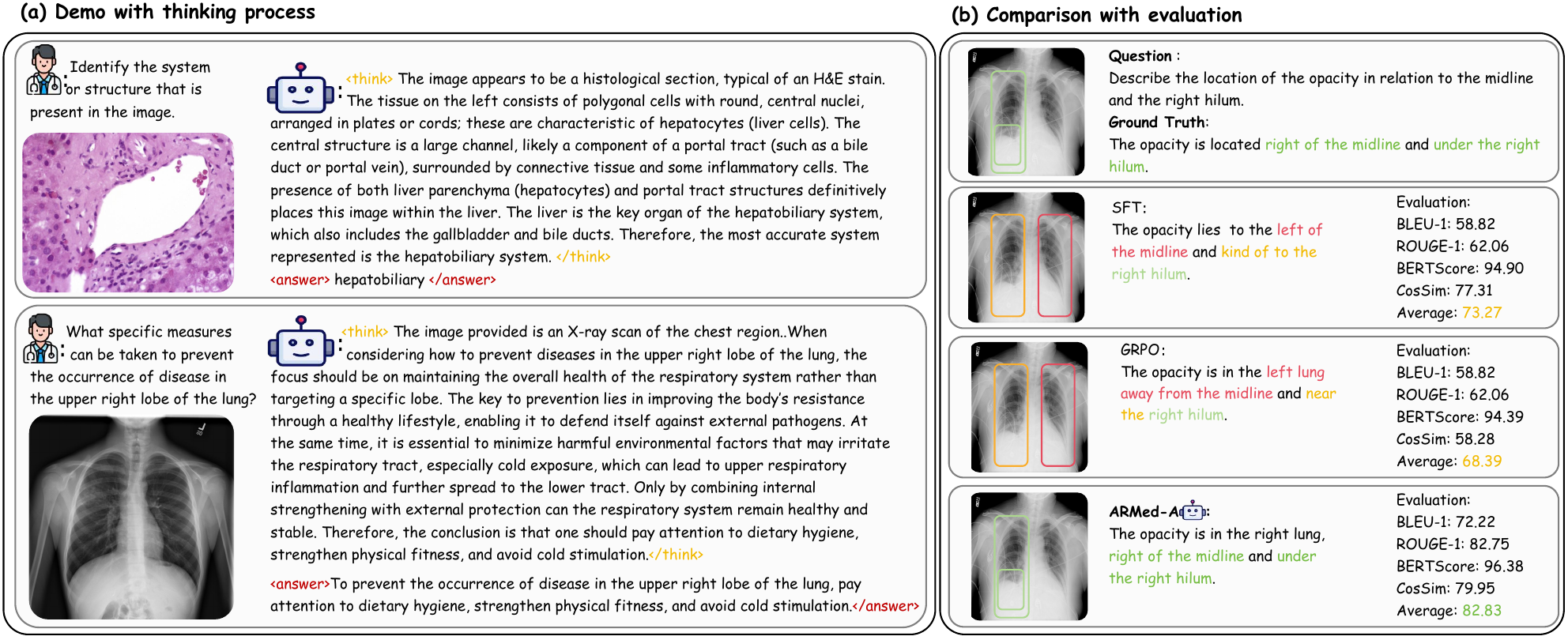}
    \caption{Demonstration of the reasoning process and evaluation across different training approaches. (a) shows  demos with reasoning process, while (b) compares model outputs and quantitative metrics from various training settings.}
    \label{fig:demo1}
    \vspace{-0.5em}
\end{figure*}

\subsection{Ablation Study}
As shown in Table~\ref{tab:qwen25_ablation} and Figure~\ref{fig:Model_Comparison}, each component of our method improves performance, with the full configuration achieving the best results across all metrics and datasets. Starting from the vanilla Qwen2.5-VL-3B baseline (46.07\%), adding textual supervision (\textit{w.\ Text}) boosts performance to 68.99\%, confirming the value of explicit text-based reasoning in medical VQA. Incorporating semantic reward modeling (\textit{w.\ Sem}) and adaptive reward adjustment (\textit{w.\ Ada}) further enhances comprehension and balance between precision and generalization, yielding additional gains.

Data augmentation (\textit{w.\ Aug}) alone improves performance to 65.63\%, and combining it with textual guidance raises the score to 72.05\%, showing their complementarity. Integrating all components achieves the highest overall average of 73.72\%, demonstrating that each module contributes uniquely and their synergy results in a robust, semantically aligned medical VQA system.

\subsection{Reward Collapse Mitigation}

From Figure~\ref{fig:Reward_Collapse}, both BERTScore and cosine similarity rewards exhibit high means and low variances. Specifically, BERTScore has a mean of 0.986 and variance of 0.029, while cosine similarity shows a mean of 0.903 and variance of 0.045. These narrowly distributed scores indicate the reward model struggles to distinguish responses of varying quality. This low variance can cause reward collapse during reinforcement learning, where the agent lacks sufficient gradient signals to optimize behavior.

In contrast, our adaptive semantic reward significantly increases variance, enhancing discriminative power. In this setting, BERTScore drops to a mean of 0.648 with a variance of 0.104, and cosine similarity to a mean of 0.514 with a variance of 0.112. This broader distribution enables more effective learning by providing richer feedback. As shown in Table~\ref{tab:qwen25_ablation} and Figure~\ref{fig:Model_Comparison}, ARMed-I, enhanced with adaptive semantic rewards, consistently outperforms Qwen2.5-VL-3B trained with GRPO across both in-domain and out-of-domain evaluations. The results demonstrate that dynamic reward modeling improves signal quality and strengthens policy robustness for medical multi-modal QA.

\subsection{Quantification Results}
As shown in Figure~\ref{fig:demo1}(a), we present the quantitative results of the visualized outputs compared with other models. It can be observed that our model produces more semantically consistent and clinically reasonable responses across various medical question types. In Figure~\ref{fig:demo1}(b), we further compare models trained under different strategies, showing that our alignment-based training approach achieves higher overall consistency and better semantic coherence between the reasoning process and the final answer.

However, it is worth noting that the current quantitative evaluation metrics are not entirely fair for open-ended medical question answering tasks. 
These metrics (\textit{e.g.}BLEU, ROUGE, BERTScore) fail to fully and truly capture the nuances of semantic correctness and deeply clinical reasoning. 
This limitation further motivates our work to explore a more semantically aware and clinically grounded evaluation metric for assessing open-ended medical VQA performance more effectively.

\section{Conclusion and Future Work}



This work presents Adaptive Reinforcement for Medical Reasoning (ARMed), a reinforcement learning framework for medical visual question answering (VQA). 
ARMed integrates domain-specific diagnostic knowledge via chain-of-thought supervision and optimizes reasoning through Group Relative Policy Optimization guided by an adaptive, clinically semantic reward. 
To overcome limitations of static similarity metrics, we design a dynamic reward mechanism that scales robustly with inter-sample variance, enhancing semantic discriminability, stabilizing policy optimization, and mitigating reward collapse. 
This adaptive formulation enables more interpretable and clinically aligned reasoning.

Extensive experiments on diverse and comprehensive medical VQA benchmarks show significant improvements in reasoning consistency, factual accuracy, and out-of-domain generalization over both supervised and reinforcement learning baselines. 
Ablation studies confirm the effectiveness of adaptive reward calibration and medical knowledge injection in reducing bias and improving robustness. 
Looking forward, we plan to extend ARMed to multi-turn and practical report-level reasoning, integrate it within clinical workflows, and develop more semantically grounded evaluation metrics for assessing open-ended medical reasoning quality. 
We discuss limitations and ethics in the \textbf{Appendix}.

{
    \small
    \bibliographystyle{ieeenat_fullname}
    \bibliography{main}
}

\clearpage
\setcounter{page}{1}
\maketitlesupplementary

\appendix
\setcounter{equation}{0}
\setcounter{figure}{0}
\setcounter{table}{0}
\renewcommand{\thefigure}{S\arabic{figure}}
\renewcommand{\thetable}{S\arabic{table}}
\renewcommand{\theequation}{S\arabic{equation}}
\renewcommand{\contentsname}{Appendix Contents}

\tableofcontents

\addtocontents{toc}{\protect\setcounter{tocdepth}{2}}
\addtocontents{toc}{\protect\setlength{\protect\cftsecnumwidth}{1.8em}}

\vspace{2em} 

\section{Limitations and Future Work}
\label{sec:appendix_limitations_future_work}

\vspace{2pt}\noindent\textbf{Scope of Evaluation.}\hspace{1ex}
Our current evaluation is limited to publicly available medical VQA benchmarks. While these datasets encompass diverse modalities and question types, they may not fully reflect the complexity and heterogeneity of real-world clinical scenarios. In future work, we plan to extend the evaluation to larger-scale and more diverse clinical datasets, including proprietary hospital archives and cross-institutional benchmarks, to better assess generalizability.

\vspace{2pt}\noindent\textbf{Model Scale and Efficiency.}\hspace{1ex}
We adopt a 3B-parameter backbone to ensure reproducibility and computational efficiency. Although this scale sufficiently demonstrates the effectiveness of our adaptive reinforcement learning framework, it may limit the upper bound of reasoning capability. Future research will explore scaling to larger model backbones and investigating lightweight distillation strategies to balance performance and efficiency in deployment.

\vspace{2pt}\noindent\textbf{Reward Design Generality.}\hspace{1ex}
The adaptive semantic reward in ARMed is primarily designed for medical reasoning tasks. While the formulation is conceptually general, its applicability to other multimodal or non-medical reasoning domains remains untested. Future directions include extending ARMed to broader visual reasoning tasks such as scientific VQA, document understanding, or instructional vision-language reasoning.

\vspace{2pt}\noindent\textbf{Evaluation Metrics.}\hspace{1ex}
We rely mainly on conventional text-based metrics (BLEU, ROUGE, BERTScore, and CosSim) to evaluate open-ended answers. These automatic measures may not adequately capture the depth, correctness, or interpretability of clinical reasoning. Future work will involve designing more semantically grounded, human-aligned, and task-specific evaluation protocols—potentially integrating expert review or reasoning trace analysis.

\section{Ethical Statement}
\label{sec:appendix_ethics}

This research strictly adheres to the ethical standards and best practices for medical AI research.

\vspace{2pt}\noindent\textbf{Data Usage and Privacy.}\hspace{1ex}
All datasets used in this work (VQA-RAD, SLAKE, PathVQA, VQA-Med, PMC-VQA, and MedXpertQA) are publicly released for academic research purposes. All data were fully anonymized and de-identified by the original providers prior to public release, containing no personally identifiable information (PII) or Protected Health Information (PHI). Our data usage fully complies with the respective dataset licenses and data use agreements. No private or clinical patient data were accessed or used in this study.

\vspace{2pt}\noindent\textbf{Intended Use and Potential Misuse.}\hspace{1ex}
Our proposed framework, \textbf{ARMed}, is designed solely for academic research on reinforcement learning and medical reasoning under open-ended visual question answering (VQA) settings. It is not intended for any direct clinical diagnosis or patient care. Although the model demonstrates improved reasoning interpretability and factual accuracy, it is a research prototype that should only be used for benchmarking, algorithmic analysis, and educational purposes. Any deployment in clinical environments requires additional regulatory validation and expert oversight. We explicitly discourage any misuse such as generating synthetic clinical advice, manipulating diagnostic workflows, or producing deceptive “medical deepfakes.”

\vspace{2pt}\noindent\textbf{Algorithmic Fairness and Bias.}\hspace{1ex}
ARMed’s performance is influenced by the quality and balance of its training datasets. Despite including multiple benchmark datasets to increase diversity, potential biases may still exist—for instance, in demographic, imaging modality, or disease-type distributions. These biases could be inadvertently learned and propagated by the model. We recommend that future work systematically investigate fairness, demographic generalization, and the ethical implications of reinforcement learning signals in medical reasoning systems.

\vspace{2pt}\noindent\textbf{Societal Impact.}\hspace{1ex}
By enhancing the interpretability and reliability of medical vision-language reasoning, this research aims to advance transparent and accountable medical AI. However, we recognize that such technologies could also amplify systemic biases or be misused if applied irresponsibly. We therefore emphasize the importance of human-in-the-loop oversight, transparent evaluation, and continued ethical scrutiny throughout model development and deployment.

\begin{figure}[!htbp]
    \centering
    \includegraphics[width=\linewidth]{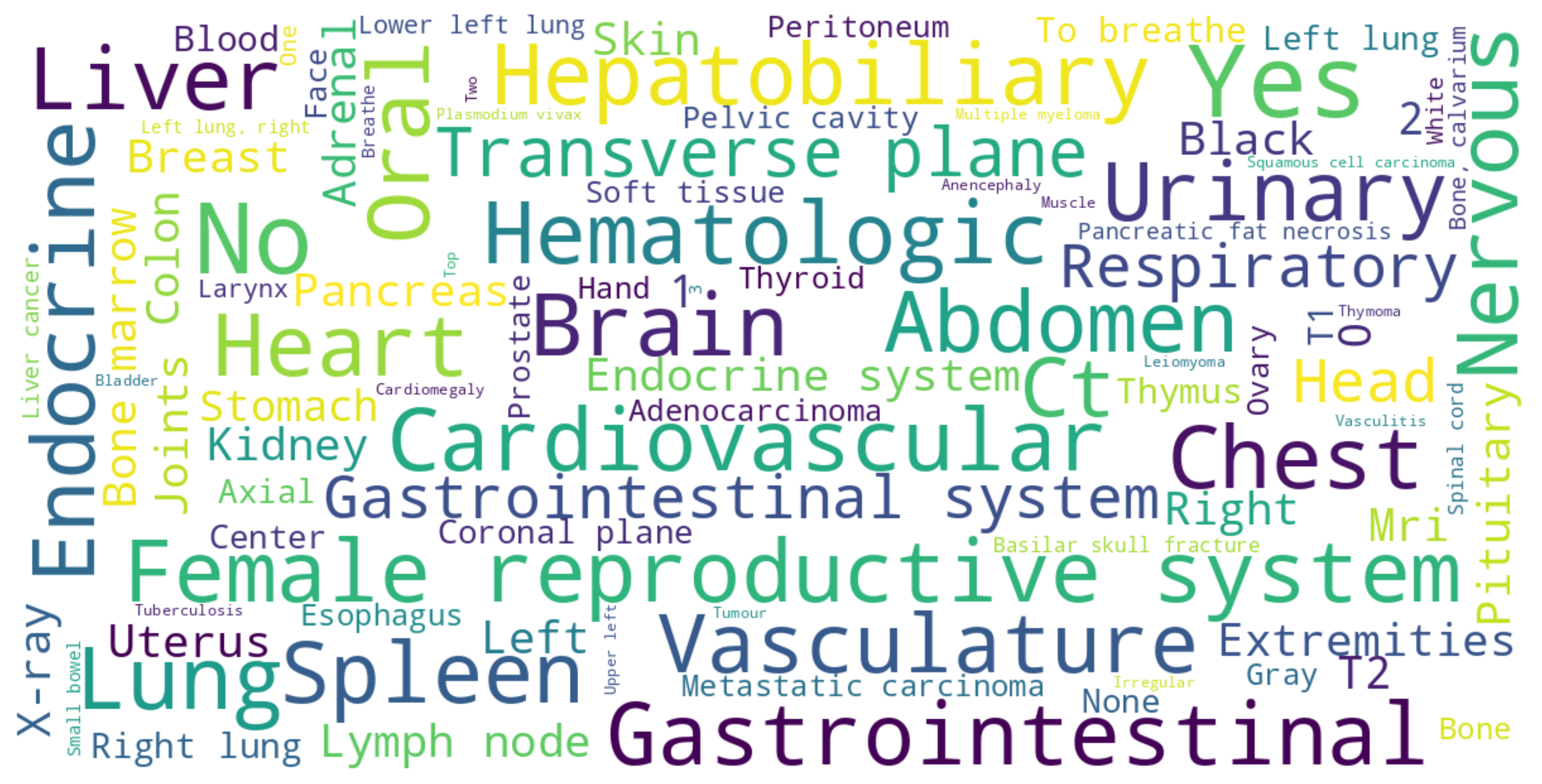}
    \caption{Word cloud of answers corresponding to high-frequency questions. The visualization highlights common medical concepts and reasoning patterns emphasized in frequently asked questions.}
    \label{fig:freq_wordcloud}
\end{figure}

\begin{figure}[!htbp]
    \centering
    \includegraphics[width=\linewidth]{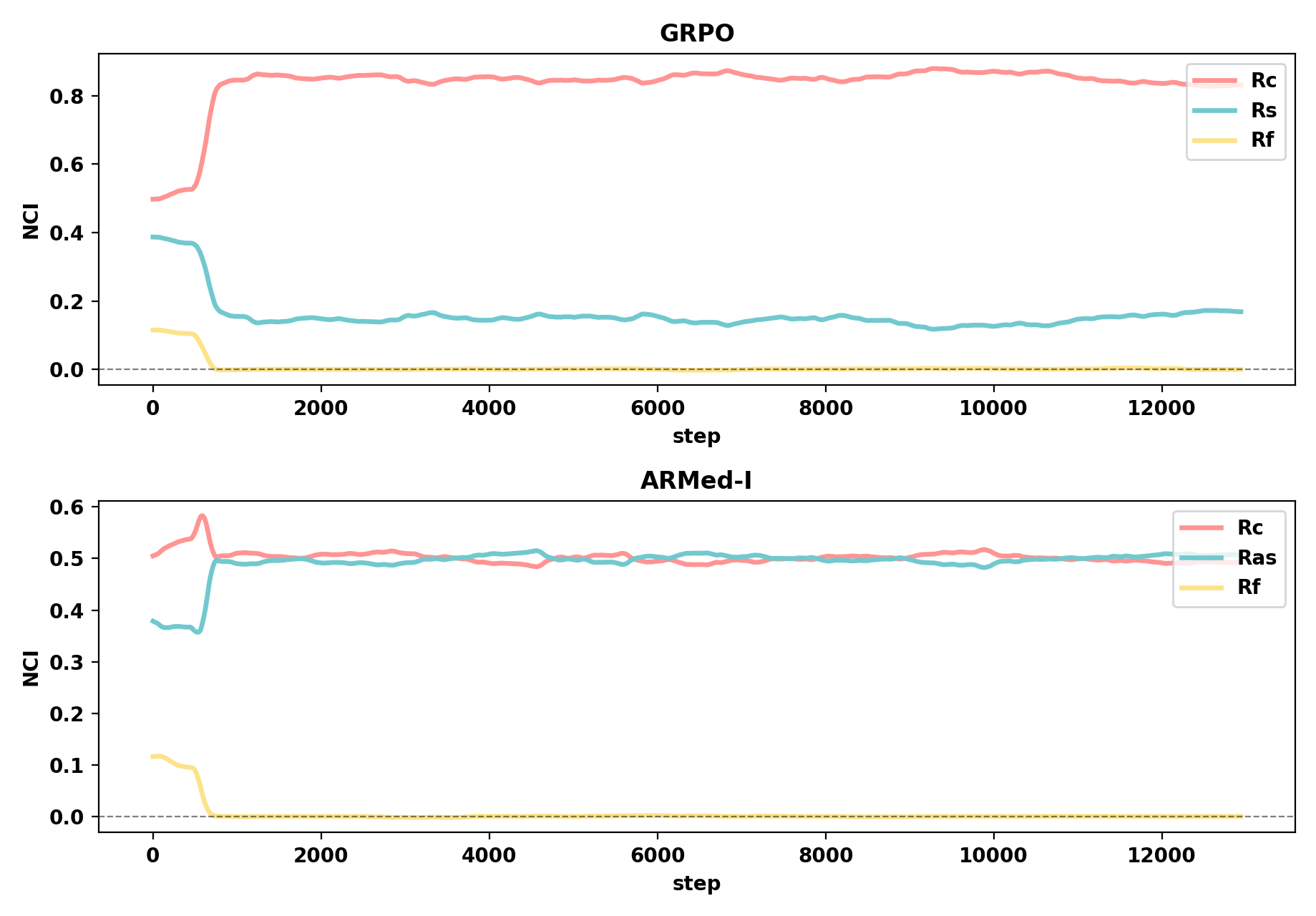}
    \caption{Comparison of NCI trajectories during training between the conventional GRPO method and our proposed ARMed-I. Rc, Rs (or Ras), and Rf denote the textual, (static/adaptive) semantic alignment, and format rewards, respectively. ARMed-I maintains more balanced and stable NCI values across reward components throughout training.}
    \label{fig:combined_nci_linesr}
\end{figure}

\section{Dataset Detailed Analysis}
\label{sec:appendix_dataset}

\subsection{Overview of Datasets}
\vspace{2pt}\noindent\textbf{PathVQA.}\hspace{1ex}
The PathVQA~\cite{he2020pathvqa} dataset is designed for pathology-related visual question answering. It encompasses a diverse range of medical images, including pathological slides, cellular microscopy, and disease-related natural images. The dataset contains 19,654 training samples and 6,719 test samples. Among the training questions, 9,903 are open-ended and 9,751 are close-ended; the test set includes 3,357 open-ended and 3,362 close-ended questions. The close-ended questions are restricted to binary (e.g., yes/no) or short categorical responses, without multiple-choice formats.

\vspace{2pt}\noindent\textbf{SLAKE.}\hspace{1ex}
The SLAKE~\cite{liu2021slake} dataset is a bilingual English–Chinese medical visual question answering benchmark. In this work, only the English portion is used. After filtering out the Chinese-language entries, the dataset comprises 4,919 training samples and 1,061 test samples. The training set includes 3,238 open-ended and 1,681 close-ended questions, while the test set contains 706 open-ended and 355 close-ended questions. The close-ended questions are limited to binary (e.g., yes/no) or short categorical responses, and do not include multiple-choice formats.

\vspace{2pt}\noindent\textbf{VQA-RAD.}\hspace{1ex}
The VQA-RAD~\cite{lau2018dataset} dataset focuses on radiology-related visual question answering. It consists of 2,244 question–answer (QA) pairs, with 1,793 designated for training and 451 for testing. The questions are categorized into open-ended and close-ended types. The training set includes 853 open-ended and 940 close-ended questions, while the test set contains 200 open-ended and 251 close-ended questions. The close-ended questions are restricted to binary (e.g., yes/no) or other short categorical responses, without multiple-choice formats.

\vspace{2pt}\noindent\textbf{VQA-Med.}\hspace{1ex}
The VQA-Med~\cite{ben2019vqa} dataset is a large-scale benchmark for multimodal understanding in biomedical imaging. It incorporates images from diverse modalities such as radiography, CT, MRI, and pathology. The dataset contains 12,792 training samples and 2,000 validation samples. The validation set includes 1,798 open-ended and 202 close-ended questions. The close-ended questions are restricted to binary (e.g., yes/no) or other short categorical responses, and do not include multiple-choice formats.

\vspace{2pt}\noindent\textbf{PMC-VQA.}\hspace{1ex}
The PMC-VQA~\cite{zhang2023pmc} dataset is a large-scale medical VQA dataset constructed from biomedical figures and captions in the PubMed Central Open Access subset. It comprises approximately 227,000 QA pairs associated with around 149,000 medical images, covering a broad spectrum of diseases and imaging modalities such as radiology, histopathology, and clinical illustrations. In this work, we adopt a manually curated, high-quality subset of 2,000 test samples for evaluation. All questions in this dataset follow a multiple-choice format.

\vspace{2pt}\noindent\textbf{MedXpertQA.}\hspace{1ex}
The MedXpertQA~\cite{zuo2025medxpertqa} dataset is a recent benchmark designed to evaluate expert-level medical reasoning in large language and vision–language models. It includes both a pure-text QA set and a multimodal subset; we use only the latter in our experiments. The multimodal portion contains 2,000 multiple-choice questions paired with medical images, covering complex diagnostic and interpretative tasks encountered in clinical practice.

\subsection{Data Preprocessing and Quality Control}
In open-ended medical VQA datasets, a common issue arises from the semantic mismatch between questions and answers, often stemming from vague or underspecified questions. This granularity inconsistency leads to cases where multiple semantically distinct answers are all valid, but only under different implicit interpretations of the same question. Such ambiguity not only complicates answer evaluation but also destabilizes policy learning in reinforcement-based training paradigms, where reward signals are sensitive to subtle semantic discrepancies.

To enhance the consistency of open-ended medical QA pairs, we take conceptual inspiration from prior work on VQA consistency auditing~\cite{rui2025improving}, which conducted refinement for both closed- and open-ended questions. In contrast, our approach focuses exclusively on open-ended questions, and we refine and extend the auditing details to better support the redefinition process tailored for reinforcement learning. This targeted design allows for a more nuanced and context-aware QA refinement strategy, as illustrated in Figure~\ref{fig:vqa_auditor_prompt}:
\begin{enumerate}
    \item \textbf{Semantic Coverage Assessment}: ensuring that each question captures the full semantic scope of its corresponding answer;
    \item \textbf{Preservation of Expressiveness}: retaining free-form, descriptive phrasing that supports nuanced and clinically relevant reasoning;
    \item \textbf{Specificity Calibration}: aligning the granularity of the question with the level of detail in the answer to avoid under- or over-generalization.
\end{enumerate}

\begin{table*}[!htbp]
\caption{Effect of the refine process on model performance across Path-VQA, SLAKE, and PMC VQA datasets.
Results are shown for SFT, GRPO, and ARMed-I with and without refinement.
\setlength{\fboxsep}{2pt}\colorbox{gray!20}{Gray-shaded} columns indicate dataset-wise averages.}
\setlength{\tabcolsep}{3pt}
\centering
\resizebox{\linewidth}{!}{
\begin{tabular}{@{}l c *{5}{c} *{5}{c} c@{\hspace{\tabcolsep}}}
\toprule
\multirow{3}{*}{Method} & \multirow{3}{*}{Refine} & \multicolumn{5}{c}{Path-VQA} & \multicolumn{5}{c}{SLAKE} & PMC VQA \\
\cmidrule(lr){3-7}\cmidrule(lr){8-12}
 &  & BLEU-1 & ROUGE-1 & BERTScore & CosSim & Avg. & BLEU-1 & ROUGE-1 & BERTScore & CosSim & Avg. & (Acc.) \\
\midrule
\multirow{2}{*}{SFT} 
 & \xmark & 48.28 & 48.94 & 96.74 & 90.98 & \cellcolor{gray!20}71.24 & 74.00 & 75.00 & 99.13 & 97.49 & \cellcolor{gray!20}86.41 & 37.55 \\
 & \cmark & 55.43 & 56.62 & 97.74 & 94.36 & \cellcolor{gray!20}76.04 & 70.58 & 71.54 & 99.10 & 97.67 & \cellcolor{gray!20}84.72 & 46.80 \\
\multirow{2}{*}{GRPO} 
 & \xmark & 42.37 & 43.51 & 97.63 & 92.32 & \cellcolor{gray!20}68.96 & 64.49 & 66.50 & 99.08 & 97.30 & \cellcolor{gray!20}81.84 & 39.55 \\
 & \cmark & 52.36 & 54.28 & 98.04 & 95.60 & \cellcolor{gray!20}75.07 & 68.30 & 69.49 & 99.13 & 97.89 & \cellcolor{gray!20}83.70 & 48.10 \\
\multirow{2}{*}{ARMed-I } 
 & \xmark & 44.76 & 46.64 & 95.25 & 89.09 & \cellcolor{gray!20}68.94 & 68.86 & 70.62 & 98.72 & 97.17 & \cellcolor{gray!20}83.84 & 46.70 \\
 & \cmark & 54.88 & 58.77 & 97.49 & 94.74 & \cellcolor{gray!20}76.47 & 68.44 & 70.49 & 98.91 & 97.54 & \cellcolor{gray!20}83.85 & 47.30 \\
\bottomrule
\end{tabular}
}

\label{tab:refine_ablation_1}
\end{table*}
\begin{table*}[!htbp]
\caption{Effect of the refine process on model performance across VQA-RAD, VQA-Med, and MedXpert datasets.
Results are shown for SFT, GRPO, and ARMed-I with and without refinement.
\setlength{\fboxsep}{2pt}\colorbox{gray!20}{Gray-shaded} columns indicate dataset-wise averages.}
\setlength{\tabcolsep}{3pt}
\centering
\resizebox{\linewidth}{!}{
\begin{tabular}{@{}l c *{5}{c} *{5}{c} c@{\hspace{\tabcolsep}}}
\toprule
\multirow{3}{*}{Method} & \multirow{3}{*}{Refine} & \multicolumn{5}{c}{VQA-RAD} & \multicolumn{5}{c}{VQA-Med} & MedXpert \\
\cmidrule(lr){3-7}\cmidrule(lr){8-12}
 &  & BLEU-1 & ROUGE-1 & BERTScore & CosSim & Avg. & BLEU-1 & ROUGE-1 & BERTScore & CosSim & Avg. & (Acc.) \\
\midrule
\multirow{2}{*}{SFT} 
 & \xmark & 56.60 & 58.87 & 98.13 & 95.52 & \cellcolor{gray!20}77.28 & 21.83 & 23.39 & 95.47 & 90.09 & \cellcolor{gray!20}57.70 & 16.65 \\
 & \cmark & 56.58 & 58.68 & 98.36 & 96.27 & \cellcolor{gray!20}77.47 & 22.44 & 24.55 & 95.54 & 90.11 & \cellcolor{gray!20}58.16 & 18.35 \\
\multirow{2}{*}{GRPO} 
 & \xmark & 47.26 & 49.81 & 98.44 & 96.20 & \cellcolor{gray!20}72.93 & 15.81 & 18.37 & 94.42 & 90.74 & \cellcolor{gray!20}54.84 & 17.95 \\
 & \cmark & 51.43 & 53.68 & 97.54 & 97.28 & \cellcolor{gray!20}74.98 & 19.95 & 20.85 & 96.32 & 91.71 & \cellcolor{gray!20}57.21 & 20.65 \\
\multirow{2}{*}{ARMed-I (Ours)} 
 & \xmark & 49.34 & 52.50 & 97.39 & 95.27 & \cellcolor{gray!20}73.63 & 17.01 & 19.62 & 96.41 & 92.39 & \cellcolor{gray!20}56.36 & 19.30 \\
 & \cmark & 54.04 & 58.89 & 98.36 & 96.84 & \cellcolor{gray!20}77.03 & 18.72 & 21.48 & 95.15 & 91.28 & \cellcolor{gray!20}56.66 & 20.90 \\
\bottomrule
\end{tabular}
}

\label{tab:refine_ablation_2}
\end{table*}

\begin{table*}[!htbp]
\caption{Ablation study results on Path-VQA, SLAKE, and PMC VQA datasets. 
\textbf{Aug} denotes \textit{augmentation}, 
\textbf{Text} denotes \textit{text reward}, 
\textbf{Sem} denotes \textit{semantic reward}, 
and \textbf{Ada} denotes \textit{adaptive semantic reward}. 
Complete results on all test sets can be found in the Appendix. 
For each metric column, \textbf{bold} denotes the best performance and \underline{underline} denotes the second-best. 
\setlength{\fboxsep}{2pt}\colorbox{gray!20}{Gray-shaded} columns correspond to dataset-specific averages.}
\setlength{\tabcolsep}{3pt}
\centering
\resizebox{\linewidth}{!}{
\begin{tabular}{@{}l*{4}{c}*{5}{c}*{5}{c}c@{\hspace{\tabcolsep}}}
\toprule
\multirow{3}{*}{Model} & \multicolumn{4}{c}{Method} & \multicolumn{5}{c}{Path-VQA} & \multicolumn{5}{c}{SLAKE} & PMC VQA \\
\cmidrule(lr){2-5}\cmidrule(lr){6-10}\cmidrule(lr){11-15}
 & \textit{w}.\ Aug & \textit{w}.\ Text & \textit{w}.\ Sem  & \textit{w}.\ Ada & BLEU-1 & ROUGE-1 & BERTScore & CosSim & Avg. & BLEU-1 & ROUGE-1 & BERTScore & CosSim & Avg. & (Acc.) \\
\midrule
\multirow{8}{*}{Qwen2.5-VL-3B}
& \xmark & \xmark & \xmark & \xmark & 5.30 & 8.20 & 86.91 & 85.10 & \cellcolor{gray!20}46.38 & 13.34 & 19.09 & 90.35 & 89.14 & \cellcolor{gray!20}52.98 & 45.70 \\
& \xmark & \cmark & \xmark & \xmark & 49.62 & 56.07 & 96.21 & 93.80 & \cellcolor{gray!20}73.93 & 64.91 & 68.76 & 98.48 & 97.19 & \cellcolor{gray!20}82.34 & 46.40 \\
& \xmark & \cmark & \cmark & \xmark & 52.36 & 54.28 & \underline{98.04} & 95.60 & \cellcolor{gray!20}75.07 & 68.30 & 69.49 & 99.13 & 97.89 & \cellcolor{gray!20}83.70 & \underline{48.10} \\
& \xmark & \cmark & \xmark & \cmark & 54.88 & 58.77 & 97.49 & 94.74 & \cellcolor{gray!20}76.47 & 68.44 & 70.49 & 98.91 & 97.54 & \cellcolor{gray!20}83.85 & 47.30 \\
& \cmark & \xmark & \xmark & \xmark & 40.72 & 43.06 & 96.49 & 94.11 & \cellcolor{gray!20}68.60 & 59.39 & 60.54 & 97.76 & 97.32 & \cellcolor{gray!20}78.75 & 43.40 \\
& \cmark & \cmark & \xmark & \xmark & \underline{61.72} & 61.92 & 97.65 & 95.74 & \cellcolor{gray!20}79.26 & 73.16 & \underline{74.87} & 98.85 & 97.15 & \cellcolor{gray!20}86.01 & 46.60 \\
& \cmark & \cmark & \cmark & \xmark & 61.23 & \underline{62.53} & 97.64 & \textbf{96.03} & \cellcolor{gray!20}\underline{79.36} & \underline{73.22} & 73.96 & \underline{99.15} & \underline{97.97} & \cellcolor{gray!20}\underline{86.08} & 48.05 \\
& \cmark & \cmark & \xmark & \cmark & \textbf{63.61} & \textbf{64.96} & \textbf{98.46} & \underline{95.81} & \cellcolor{gray!20}\textbf{80.71} & \textbf{76.14} & \textbf{76.65} & \textbf{99.46} & \textbf{98.13} & \cellcolor{gray!20}\textbf{87.60} & \textbf{48.75} \\
\bottomrule
\end{tabular}
}

\label{tab:ablation_section_1}
\end{table*}

\begin{table*}[!htbp]
\caption{Ablation study results on VQA-RAD, VQA-Med, and MedXpertQA datasets. 
\textbf{Aug} denotes \textit{augmentation}, 
\textbf{Text} denotes \textit{text reward}, 
\textbf{Sem} denotes \textit{semantic reward}, 
and \textbf{Ada} denotes \textit{adaptive semantic reward}. 
For each metric column, \textbf{bold} denotes the best performance and \underline{underline} denotes the second-best. 
\setlength{\fboxsep}{2pt}\colorbox{gray!20}{Gray-shaded} columns correspond to dataset-specific averages.}
\setlength{\tabcolsep}{3pt}
\centering
\resizebox{\linewidth}{!}{
\begin{tabular}{@{}l*{4}{c}*{5}{c}*{5}{c}c@{\hspace{\tabcolsep}}}
\toprule
\multirow{3}{*}{Model} & \multicolumn{4}{c}{Method} & \multicolumn{5}{c}{VQA-RAD} & \multicolumn{5}{c}{VQA-Med} & MedXpert \\
\cmidrule(lr){2-5}\cmidrule(lr){6-10}\cmidrule(lr){11-15}
 & \textit{w}.\ Aug & \textit{w}.\ Text & \textit{w}.\ Sem  & \textit{w}.\ Ada & BLEU-1 & ROUGE-1 & BERTScore & CosSim & Avg. & BLEU-1 & ROUGE-1 & BERTScore & CosSim & Avg. & (Acc.) \\
\midrule
\multirow{8}{*}{Qwen2.5-VL-3B}
& \xmark & \xmark & \xmark & \xmark & 9.58 & 13.38 & 88.46 & 87.64 & \cellcolor{gray!20}49.77 & 4.88 & 6.39 & 69.47 & 67.01 & \cellcolor{gray!20}36.94 & 11.60 \\
& \xmark & \cmark & \xmark & \xmark & 52.55 & 59.18 & 97.84 & 96.48 & \cellcolor{gray!20}76.51 & 13.90 & 17.39 & 94.53 & 90.96 & \cellcolor{gray!20}54.20 & 19.90 \\
& \xmark & \cmark & \cmark & \xmark & 51.43 & 53.68 & 97.54 & \underline{97.28} & \cellcolor{gray!20}74.98 & 19.95 & 20.85 & \underline{96.32} & 91.71 & \cellcolor{gray!20}57.21 & 20.65 \\
& \xmark & \cmark & \xmark & \cmark & 54.04 & 58.89 & 98.36 & 96.84 & \cellcolor{gray!20}77.03 & 18.72 & 21.48 & 95.15 & 91.28 & \cellcolor{gray!20}56.66 & 20.90 \\
& \cmark & \xmark & \xmark & \xmark & 49.61 & 51.53 & 97.62 & 96.93 & \cellcolor{gray!20}73.92 & 20.52 & \underline{22.50} & 95.29 & 92.04 & \cellcolor{gray!20}57.59 & 19.25 \\
& \cmark & \cmark & \xmark & \xmark & 57.63 & \underline{61.09} & 98.66 & 97.23 & \cellcolor{gray!20}78.65 & 20.75 & 22.42 & 95.30 & 91.96 & \cellcolor{gray!20}57.61 & 20.50 \\
& \cmark & \cmark & \cmark & \xmark & \underline{57.78} & 60.93 & \underline{98.68} & 97.26 & \cellcolor{gray!20}\underline{78.66} & \underline{20.84} & 22.33 & 96.31 & \underline{92.34} & \cellcolor{gray!20}57.96 & \underline{20.95} \\
& \cmark & \cmark & \xmark & \cmark & \textbf{58.72} & \textbf{61.10} & \textbf{98.69} & \textbf{97.29} & \cellcolor{gray!20}\textbf{78.95} & \textbf{21.51} & \textbf{23.17} & \textbf{96.56} & \textbf{92.36} & \cellcolor{gray!20}\textbf{58.40} & \textbf{22.30} \\
\bottomrule
\end{tabular}
}
\label{tab:ablation_section2}
\end{table*}

These refinements, performed as a preprocessing step, enhance question-answer alignment and improve semantic fidelity, thereby providing a more stable foundation for policy optimization under the GRPO framework.

To illustrate the impact of our refinement strategy, several representative examples of question rewriting are presented in Figure~\ref{fig:vqa_refinement_examples}. Each example showcases the transformation of an ambiguous or underspecified original question into a more precise and semantically aligned version, while preserving its open-ended nature. These refinements enhance the alignment between the question and its corresponding answer, thereby reducing potential mismatches that could mislead reinforcement-based policy updates. Specifically, the revisions clarify vague phrasing, increase semantic coverage, and calibrate specificity to match the expert-provided answers. This process serves as a critical preprocessing step, ensuring higher consistency and reasoning quality during training.

\begin{figure*}[!p]
\centering
\begin{tcolorbox}[
  title=VQA Refinement Prompt,
  colback=blue!3!white,
  colframe=blue!80!black,
  coltitle=white,
  colbacktitle=blue!60!black,
  fonttitle=\bfseries,
  fontupper=\normalsize,
  fontupper=\small,
  boxrule=0.6pt,
  arc=1mm,
  top=2pt, bottom=2pt, left=5pt, right=5pt,
  width=0.8\textwidth,
  sharp corners
]

\texttt{ori\_q}: \texttt{\{question\}} \\
\texttt{ori\_a}: \texttt{\{answer\}}

\vspace{0.5em}
\textbf{Role:} \textbf{“QA-Consistency Auditor”} – an expert data curator.  
Your task is to audit and refine \textbf{open-ended} and \textbf{binary (yes/no)} visual-question-answering (VQA) pairs by evaluating the alignment between the question (\texttt{ori\_q}) and the answer (\texttt{ori\_a}).  
Revise them when necessary to ensure clarity, precision, and consistency.  
Visual content is unavailable—base all decisions solely on textual information.

\vspace{0.5em}
\textbf{Process:}
\begin{enumerate}[label=\arabic*, itemsep=0.3em, topsep=0.3em]
    \item Parse the original question (\texttt{ori\_q}) and original answer (\texttt{ori\_a}).
    \item Based solely on linguistic content, infer the most likely intended meaning (\texttt{Expert\_Guess}).
    \item Compare \texttt{Expert\_Guess} with \texttt{ori\_a} for consistency in scope, granularity, semantic focus, and answerability.
    \item Assign a status from the following:
    \begin{itemize}[itemsep=0pt]
        \item \textbf{consistent}: The question already elicits exactly the content and structure found in the answer.
        \item \textbf{needs\_fix}: The question is underspecified, ambiguous, or mismatched with the answer.
        \item \textbf{drop}: The QA pair is incoherent, contradictory, or cannot be repaired.
    \end{itemize}
\end{enumerate}

\vspace{0.5em}
\textbf{Guidelines for Fixing (if \texttt{status = "needs\_fix"}):}

General Constraints:
\begin{itemize}[itemsep=0pt]
    \item Ensure the revised question fully supports and specifies all semantic information in the answer.
    \item Maintain alignment in scope, granularity, structure, and ordering between question and answer.
    \item Preserve the form of the answer (e.g., short phrase, description, yes/no).
\end{itemize}

For Open-Ended Questions:
\begin{itemize}[itemsep=0pt]
    \item Begin with a directive verb (e.g., ``Describe'', ``Explain'', ``Identify'', ``What is'', ``Why is'').
    \item Explicitly request each semantic element contained in the answer.
    \item Match the level of detail; avoid over-generalization or unnecessary specificity.
    \item Keep the ordering of components consistent between question and answer.
    \item Avoid vague references such as ``this'', ``that'', ``there'', or ``it'' unless the referent is clear from text alone.
\end{itemize}

For Yes/No Questions:
\begin{itemize}[itemsep=0pt]
    \item Ensure the question expresses a clear, testable proposition that can be answered by ``yes'' or ``no''.
    \item Explicitly include the subject and the attribute or action being confirmed or denied.
    \item Remove ambiguity and implicit assumptions (e.g., clarify ``Is it raining?'' to ``Is the sky raining in this scene?'' if needed).
    \item Do not convert yes/no answers into descriptive ones or vice versa.
    \item Maintain grammatical completeness and clarity.
\end{itemize}

\vspace{0.5em}
\textbf{Output Format:}

Return exactly \textbf{one JSON object} with the schema:
\begin{verbatim}
{
  "status": "consistent | needs_fix | drop",
  "ori_q": "<string>",
  "ori_a": "<string>",
  "new_q": "<string>",
  "new_a": "<string>",
  "notes": "rationale for decision"
}
\end{verbatim}
Do not return anything other than the JSON object.

\end{tcolorbox}
\caption{Prompt template for open-ended and binary (yes/no) VQA refinement.}
\label{fig:vqa_auditor_prompt}
\end{figure*}

\begin{figure*}[htbp]
\centering
\begin{tcolorbox}[
  colback=gray!3,
  colframe=gray!60!black,
  title=Examples of VQA Refinement,
  boxrule=0.5pt,
  arc=2mm,
  fonttitle=\bfseries,
  fontupper=\normalsize,
  left=2mm,
  right=2mm,
  top=1mm,
  bottom=1mm,
  width=0.95\linewidth
]

\textbf{Example 1} \\
\texttt{Original Question:} Where does the trabecular bone forming the marrow space show? \\
\texttt{Refined Question:} In what location does the trabecular bone forming the marrow space typically appear? \\
\texttt{Answer:} At the margins.

\vspace{0.8em}

\textbf{Example 2} \\
\texttt{Original Question:} What does the cortical bone forming the outer shell show? \\
\texttt{Refined Question:} Describe the structural features visible in the cortical bone forming the outer shell, including the arrangement of lamellae, presence of osteocytic lacunae, and relationship to blood vessels. \\
\texttt{Answer:} Concentric lamellae along with osteocytic lacunae surrounding central blood vessels.

\vspace{0.8em}

\textbf{Example 3} \\
\texttt{Original Question:} What does process begin as? \\
\texttt{Refined Question:}Describe how the process begins, including its initial form and location within the body. \\
\texttt{Answer:} The process begins as a focus of microabscess in a vascular loop in the marrow, which then expands to stimulate further activity.

\end{tcolorbox}
\caption{Illustration of open-ended VQA refinement examples. Each pair consists of an original question, a refined version with improved specificity and semantic clarity, and the corresponding expert-provided answer.}
\label{fig:vqa_refinement_examples}
\end{figure*}

\subsection{Performance of Data Preprocessing}
To evaluate the effectiveness of our proposed VQA Refinement framework, we conduct ablation studies on three representative methods—SFT, GRPO, and ARMed-I—across multiple medical VQA datasets. As shown in Table~\ref{tab:refine_ablation_1} and Table~\ref{tab:refine_ablation_2}, VQA Refinement consistently and significantly improves model performance under all settings.

\section{Ablation Study}
\label{sec:appendix_ablation_Study}
Due to space limitations, we only present part of the ablation study results in the main text. The complete results of the ablation experiments are provided in Table~\ref{tab:ablation_section_1} and Table~\ref{tab:ablation_section2}.

\begin{table*}[!htbp]
\centering
\caption{Hyperparameters for Adaptive Semantic Alignment Reward}
\begin{tabular}{@{}lll@{}}
\toprule
\textbf{Hyperparameter} & \textbf{Description} & \textbf{Example Value} \\
\midrule
$\rho$ & Minimum ratio for filtering valid rewards & 0.8 \\
$L_{\max}$ & Max length of historical reward buffer & 2000 \\
$p$ & Percentile to compute dynamic threshold & 0.5 \\
$\delta_{\max}$ & Max threshold change per step & 0.01 \\
$T_{\min}$ & Minimum allowed threshold value & 0.0 \\
$T_{\max}$ & Maximum allowed threshold value & 0.995 \\
$\epsilon$ & Small constant to avoid division by zero & $1 \times 10^{-8}$ \\
$\alpha_{\text{pos}}$ & Steepness of S-curve (positive side) & 5.0 \\
$\alpha_{\text{neg}}$ & Steepness of S-curve (negative side) & 2.0 \\
$\tau$ & Word frequency threshold& 5\\
\bottomrule
\end{tabular}
\label{tab:reward_hyperparams}

\end{table*}

\section{Reward Collapse}
\subsection{Normalized Contribution Index (NCI)}

\vspace{2pt}\noindent\textbf{Motivation.}\hspace{1ex}
In Group Relative Policy Optimization (GRPO), the policy update is governed by the advantage \(A_i\), obtained via group-wise normalization of the total reward:
\begin{equation}
A_i = \frac{r_i - \bar{r}}{\sigma(r)}.
\end{equation}
Let the total reward be a weighted sum of distinct components, \(r = \sum_k \gamma_k R_k\).
Substituting this expression into the advantage formulation and rearranging terms yields:
\begin{equation}
A_i = \sum_k 
\underbrace{\left[ \gamma_k \frac{\sigma(R_k)}{\sigma(r)} \right]}_{\text{Effective Coefficient}} 
\underbrace{\left( \frac{R_{k,i} - \bar{R}_k}{\sigma(R_k)} \right)}_{\text{Normalized Component}}.
\label{eq:advantage_decomp}
\end{equation}

Equation~\ref{eq:advantage_decomp} reveals a key structural insight: 
the \textit{effective optimization weight} of component \(k\) is not simply the scalar \(\gamma_k\), but is modulated by the ratio \(\frac{\sigma(R_k)}{\sigma(r)}\).
This leads to what we term \textit{Reward Collapse}---if a particular reward component (e.g., a semantic constraint) exhibits much lower variance than the overall reward (\(\sigma(R_k) \ll \sigma(r)\)), its effective coefficient approaches zero. 
Consequently, the advantage signal \(A_i\)---and thus the policy gradient---becomes insensitive to that component, rendering it effectively invisible during optimization.

\vspace{2pt}\noindent\textbf{From Advantage to Component Contribution.}\hspace{1ex}
Starting from Eq.~\ref{eq:advantage_decomp}, we can rewrite the normalized advantage as
\begin{equation}
\begin{split}
    A_i &= \sum_k \alpha_k Z_{k,i}, \\
\alpha_k &= \gamma_k \frac{\sigma(R_k)}{\sigma(r)}, \\
Z_{k,i} &= \frac{R_{k,i} - \bar{R}_k}{\sigma(R_k)}.
\end{split}
\label{eq:A-Zk-simple}
\end{equation}
Each $Z_{k,i}$ represents a standardized (zero-mean, unit-variance) version of reward component $R_k$, while $\alpha_k$ determines its effective scaling in the normalized advantage.

Since $A_i$ is normalized within each group, we have $\mathrm{Var}(A)=1$.  
Taking the covariance between $A$ and itself yields:
\begin{equation}
\begin{split}
    1 &= \mathrm{Cov}(A, A) = \mathrm{Cov}\!\left(\sum_k \alpha_k Z_k,\, A\right)\\
   &= \sum_k \alpha_k\, \mathrm{Cov}(Z_k, A).
\end{split}
\label{eq:cov-sum}
\end{equation}

\vspace{2pt}\noindent\textbf{Relating to Reward Components.}\hspace{1ex}
Because $A = (r - \bar{r}) / \sigma(r)$ is a linear normalization of $r$, the covariance scales as
\begin{equation}
\mathrm{Cov}(R_k, A) = \frac{\mathrm{Cov}(R_k, r)}{\sigma(r)}.
\end{equation}
Since linear normalization preserves correlation, it follows that
\begin{equation}
\mathrm{Cov}(Z_k, A) = \rho(Z_k, A) = \rho(R_k, A) = \rho(R_k, r).
\end{equation}
In other words, the standardized component $Z_k$ has the same correlation with the normalized advantage $A$ as $R_k$ has with the total reward $r$.

\vspace{2pt}\noindent\textbf{Definition of NCI.}\hspace{1ex}
Substituting this relation into Eq.~\ref{eq:cov-sum} gives:
\begin{equation}
1 = \sum_k \alpha_k \rho(R_k, r)
   = \sum_k \gamma_k \frac{\sigma(R_k)}{\sigma(r)} \rho(R_k, r).
\end{equation}
This motivates defining the \textbf{Normalized Contribution Index (NCI)} for each component as
\begin{equation}
\text{NCI}_k = \gamma_k \frac{\sigma(R_k)}{\sigma(r)} \rho(R_k, r),
\label{eq:nci-simple}
\end{equation}
which satisfies $\sum_k \text{NCI}_k = 1$.

\vspace{2pt}\noindent\textbf{Interpretation.}\hspace{1ex}
Each $\mathrm{NCI}_k$ quantifies the proportional contribution of component $k$ to the total variance of the normalized advantage $A$:
\begin{itemize}[leftmargin=2em]
    \item $\gamma_k$ — design weight (intended importance);
    \item $\displaystyle \frac{\sigma(R_k)}{\sigma(r)}$ — relative signal strength;
    \item $\rho(R_k, r)$ — alignment with the overall reward.
\end{itemize}
Intuitively, $\mathrm{NCI}_k$ measures how much the component $R_k$ contributes to the unit-variance signal of $A$.
When $\mathrm{NCI}_k \approx 0$, that component has effectively collapsed and no longer influences the policy update.

\subsection{Quantitative Analysis based on NCI}

From Figure~\ref{fig:combined_nci_linesr}, we can clearly observe the evolution of NCI values for each reward component during training under both GRPO and ARMed-I. 
In GRPO, the NCI of the format reward ($R_f$) rapidly drops to nearly zero, indicating that it barely contributes to the gradient update. 
The NCI of the static semantic reward ($R_s$) also gradually decreases to around 0.2, suggesting that its guidance becomes much weaker compared to the textual reward ($R_c$). 
This phenomenon is a clear manifestation of the \textit{reward collapse} problem discussed earlier. 
In contrast, under our proposed ARMed-I, the NCI values of the textual reward ($R_c$) and the adaptive semantic alignment reward ($R_{as}$) remain at comparable levels after stabilization, 
indicating that both semantic and textual signals are effectively guiding the model toward the desired optimization direction.

\section{Implement Details}
\label{sec:implement detail}
\subsection{Hyperparameters}
\vspace{2pt}\noindent\textbf{GRPO.}\hspace{1ex}
The detailed training configurations for GRPO are provided in the implementation details section in main text. Specifically, the hyperparameters used in our design of the adaptive semantic alignment reward are listed in Table~\ref{tab:reward_hyperparams}. These include thresholding criteria, buffer limits, and the shaping parameters of the reward transformation function, all of which are carefully selected to ensure both training stability and semantic discrimination.

\vspace{2pt}\noindent\textbf{Supervise Fine-Tuning.}\hspace{1ex}
For Supervised Fine-Tuning (SFT), we employ the LLaMA Factory with full-parameter tuning, using the same base model and hardware setup. The batch size is set to 16 per GPU without gradient accumulation. A cosine learning rate schedule is adopted with an initial rate of $1 \times 10^{-6}$ as same as GRPO. 

\subsection{Models}
\vspace{2pt}\noindent\textbf{BERTScore.}\hspace{1ex}
To compute the BERTScore, we adopt PubMedBERT~\cite{pubmedbert} , a transformer-based language model that is pretrained from scratch on the PubMed abstracts and full-text biomedical articles. Unlike general-domain models such as BERT-base or RoBERTa, PubMedBERT is specifically optimized for understanding biomedical terminology, abbreviations, and domain-specific sentence structures. Its pretraining corpus ensures a closer alignment with the linguistic patterns found in medical visual question answering tasks, thereby enabling more accurate semantic similarity estimation between generated answers and ground truth responses.

\vspace{2pt}\noindent\textbf{Cosine Similarity.}\hspace{1ex}
For cosine similarity evaluation, we utilize BioBERT-mnli-snli-scinli-scitail-mednli-stsb~\cite{deka2022evidence}, a domain-adapted sentence embedding model that integrates multiple biomedical natural language inference and semantic textual similarity datasets during training. This model is designed to capture nuanced relational semantics in the biomedical context, making it well-suited for reward evaluation tasks in medical reinforcement learning. By leveraging embeddings from this multi-task fine-tuned BioBERT variant, we aim to provide a robust and context-aware similarity signal that reflects medical factual consistency and reasoning coherence.

\subsection{Medical Thinking Knowledge Injection}
Our training set consists of a total of 26,366 questions. Following our selection strategy, we curated a subset of 4,581 representative questions. Among them, 940 questions were identified as high-frequency types, while the remaining 3,641 were categorized as low-frequency types. A word cloud visualization of the answers corresponding to high-frequency questions is shown in Figure~\ref{fig:freq_wordcloud}.

\section{Supplementary Quantitative Results}
\label{sec:appendix_quant}

\begin{figure*}[!htbp]
    \centering
    \includegraphics[width=\linewidth]{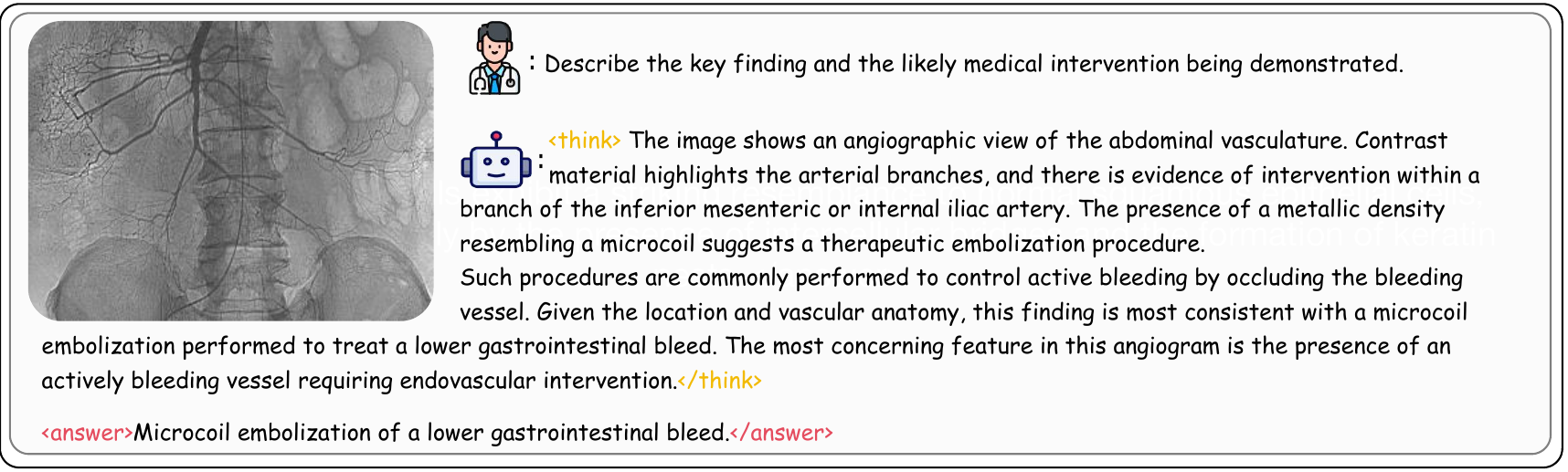}
    \caption{A quantitative result of our ARMed}
    \label{fig:demo4}
\end{figure*}

To further validate the effectiveness and interpretability of the proposed \textbf{ARMed} framework, this section presents several representative demonstration results (see Figure~\ref{fig:demo1}-\ref{fig:demo4}). These examples cover diverse medical visual question answering (VQA) scenarios, illustrating the model's reasoning capability across different modalities and question types. It can be observed that ARMed maintains both semantic consistency and clinical plausibility throughout the reasoning process. The reasoning trace (\texttt{<think>} block) and the corresponding final answer (\texttt{<answer>} block) jointly constitute a transparent and analyzable decision pathway, further confirming the stability and generalizability of the proposed method in open-ended medical reasoning tasks.

\begin{figure*}[!htbp]
    \centering
    \includegraphics[width=\linewidth]{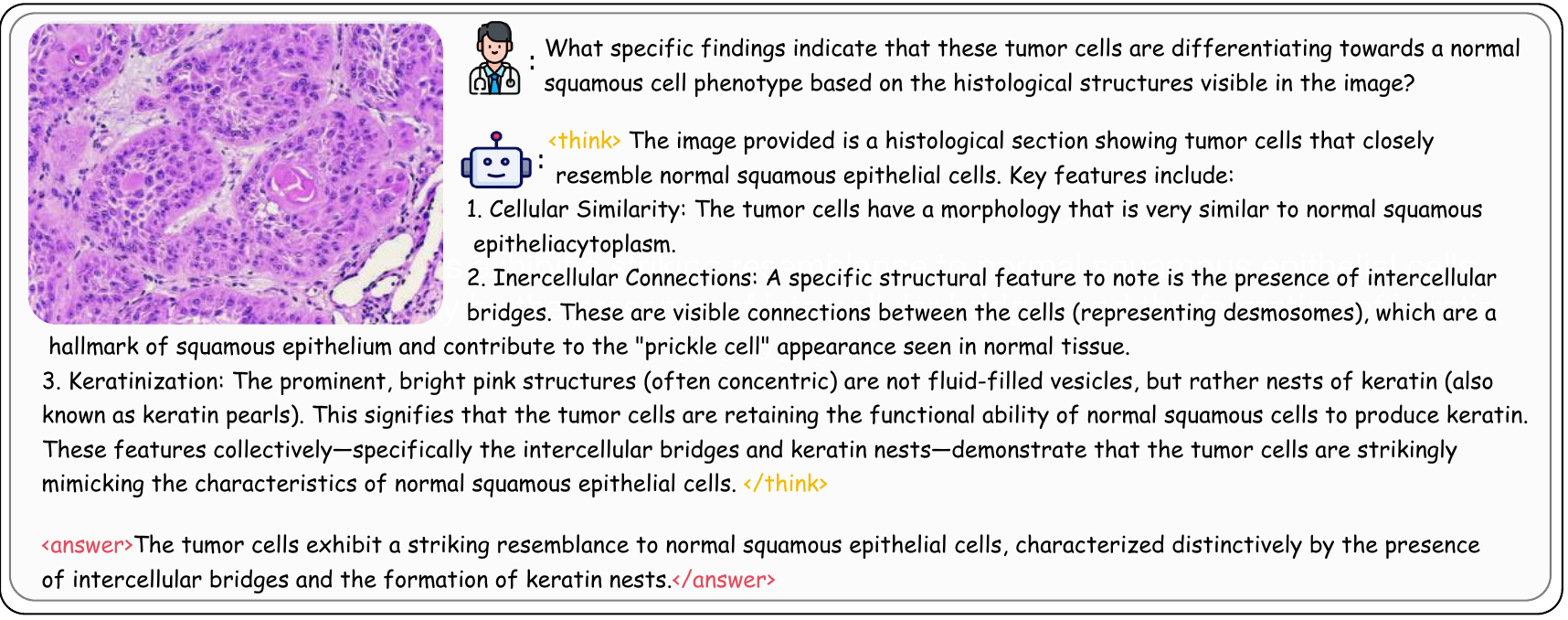}
    \caption{A quantitative result of our ARMed}
    \label{fig:demo1}
\end{figure*}

\begin{figure*}[!htbp]
    \centering
    \includegraphics[width=\linewidth]{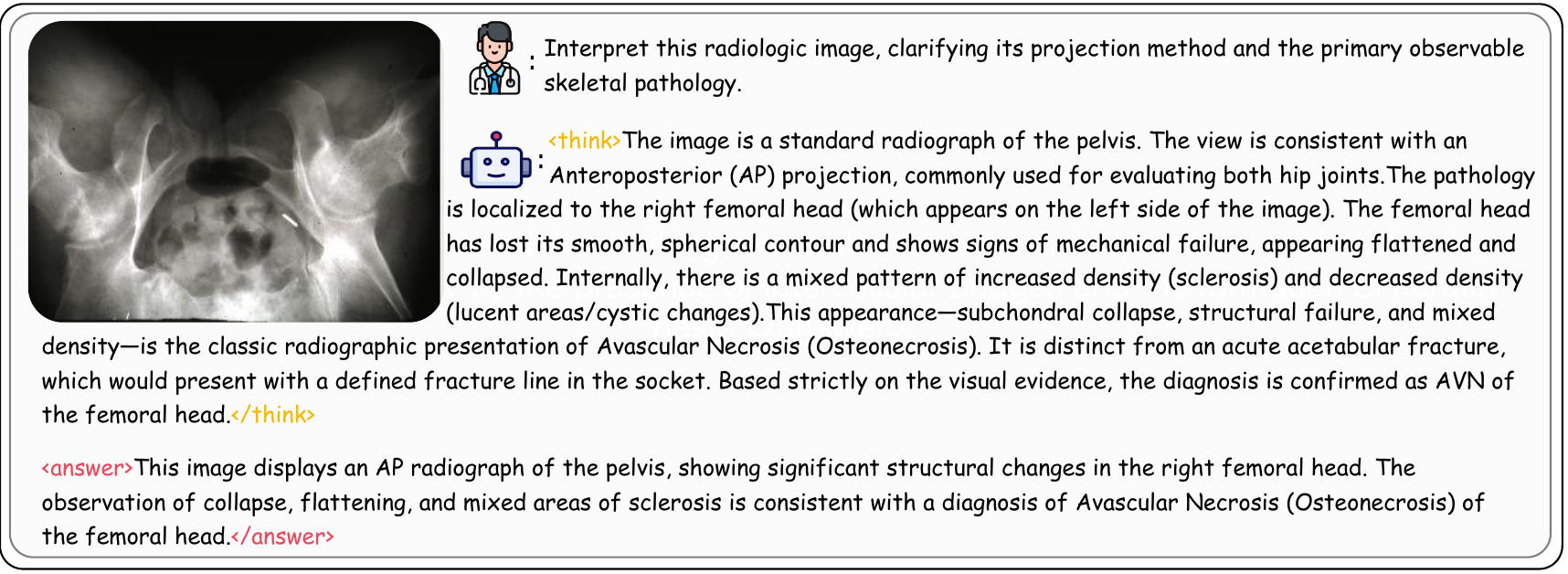}
    \caption{A quantitative result of our ARMed}
    \label{fig:demo2}
\end{figure*}

\begin{figure*}[!htbp]
    \centering
    \includegraphics[width=\linewidth]{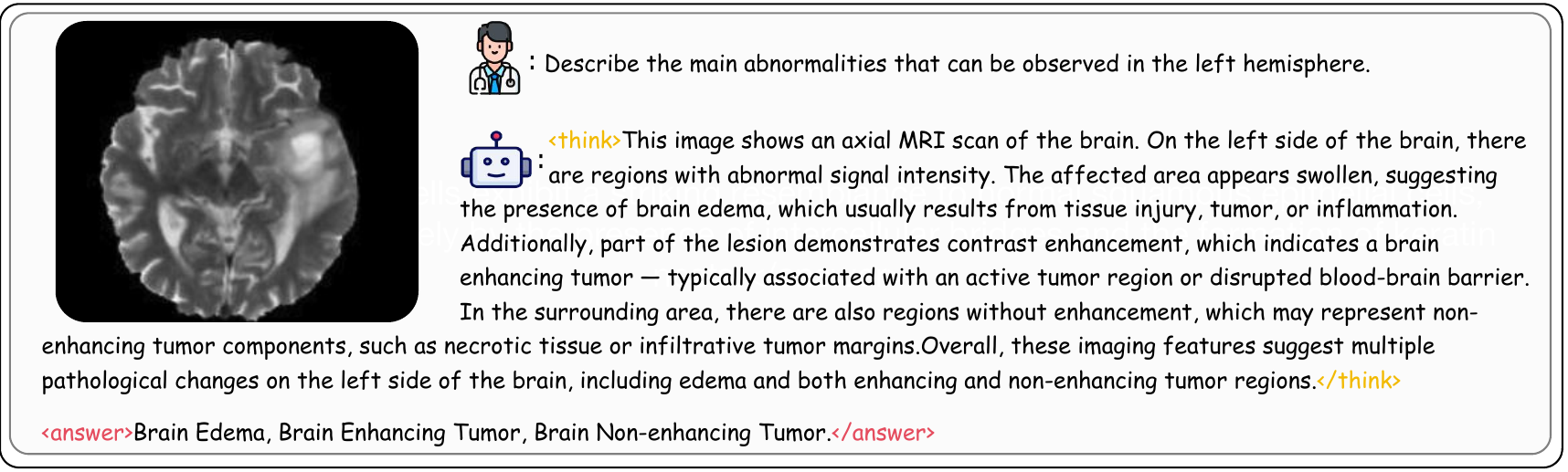}
    \caption{A quantitative result of our ARMed}
    \label{fig:demo3}
\end{figure*}

\end{document}